\documentclass[journal,transmag]{IEEEtran} 

\pdfoutput=1 

\usepackage{flushend}
\usepackage{cite}

\usepackage{color,graphicx,enumerate}
\usepackage{amsmath,amsfonts,amssymb}
\usepackage[font=small,labelfont=bf]{caption}
\usepackage{subcaption}

\newif\ifannotated

\annotatedfalse

\ifannotated
\newcommand{\add}[1]{{\color{blue}{#1}}}
\newcommand{\delete}[1]{{\color{red}{\sout{#1}}}}
\newcommand{\replace}[2]{{\delete{#1}} : {\add{#2}}}
\newcommand{\margincomment}[1]{\marginpar{$\Rightarrow$\color{red}\fbox{\parbox{\linewidth}{\color{black}\scriptsize#1}}}}
\else
\newcommand{\add}[1]{{{#1}}}
\newcommand{\delete}[1]{{\unskip \ignorespaces}}
\newcommand{\replace}[2]{{\delete{#1}}  {\add{#2}}}
\newcommand{\margincomment}[1]{}
\fi

\begin{document}
%
\title{FlatCam: Thin, Bare-Sensor Cameras using Coded Aperture and Computation}


\author{\IEEEauthorblockN{M. Salman Asif,\IEEEauthorrefmark{1}
Ali Ayremlou,\IEEEauthorrefmark{1}
Aswin Sankaranarayanan,\IEEEauthorrefmark{2}
Ashok Veeraraghavan,\IEEEauthorrefmark{1} and
Richard Baraniuk\IEEEauthorrefmark{1}}
\IEEEauthorblockA{\IEEEauthorrefmark{1}Department of Electrical and Computer Engineering, Rice University, Houston, TX 77005, USA}
\IEEEauthorblockA{\IEEEauthorrefmark{2}Department of Electrical and Computer Engineering, Carnegie Mellon University,  Pittsburgh, PA 15213, USA}
\thanks{
	Corresponding author: M. Salman Asif (email: sasif@rice.edu).
	}
}
 
\IEEEtitleabstractindextext{%
\begin{abstract}
FlatCam is a thin form-factor lensless camera that consists of a coded mask placed on top of a bare, conventional sensor array. Unlike a traditional, lens-based camera where an image of the scene is directly recorded on the sensor pixels, each pixel in FlatCam records a linear combination of light from multiple scene elements. A computational algorithm is then used to demultiplex the recorded measurements and reconstruct an image of the scene. FlatCam is an instance of a coded aperture imaging system; however, unlike the vast majority of related work, we place the coded mask extremely close to the image sensor that can enable a thin system. We employ a separable mask to ensure that both calibration and image reconstruction are scalable in terms of memory requirements and computational complexity. We demonstrate the potential of the FlatCam design using two prototypes: one at visible wavelengths and one at infrared wavelengths. 

\end{abstract}

}

\maketitle

\IEEEdisplaynontitleabstractindextext

%
\IEEEpeerreviewmaketitle

\section{Introduction}
%
A range of new imaging applications is driving the miniaturization of cameras. As a consequence, significant progress has been made towards minimizing the total volume of the camera, which has enabled new applications in endoscopy, pill cameras, and in vivo microscopy. 
Unfortunately, this strategy of miniaturization has an important shortcoming: the amount of light collected at the sensor decreases dramatically as the lens aperture and the sensor size become smaller.
As a consequence, ultra-miniature imagers built simply by scaling down the optics and sensors suffer from extremely low light collection.

In this paper, we present a camera architecture that we call FlatCam, which is inspired by coded aperture imaging principles pioneered in astronomical X-ray and gamma-ray imaging \cite{dicke1968scatter,fenimore1978coded,gottesman1989new,cannon1980coded,durrant1999application}. 
Our proposed FlatCam design uses a large photosensitive area with a very thin form factor. The FlatCam achieves thin form factor by dispensing with a lens and replacing it with a coded, binary mask placed almost immediately atop a bare conventional sensor array. 
The image formed on the sensor can be viewed as a superposition of many pinhole images. Thus, the light collection ability of such a coded aperture system is proportional to the size of the sensor and the transparent regions (pinholes) in the mask.  
In contrast, the light collection ability of a miniature, lens-based camera is limited by the lens aperture size, which is restricted by the requirements on the device thickness. 

\begin{figure*}[t]
	\centering
	\includegraphics[width=0.75\textwidth,page=1, trim = 0mm 75mm 0mm 0mm, clip]{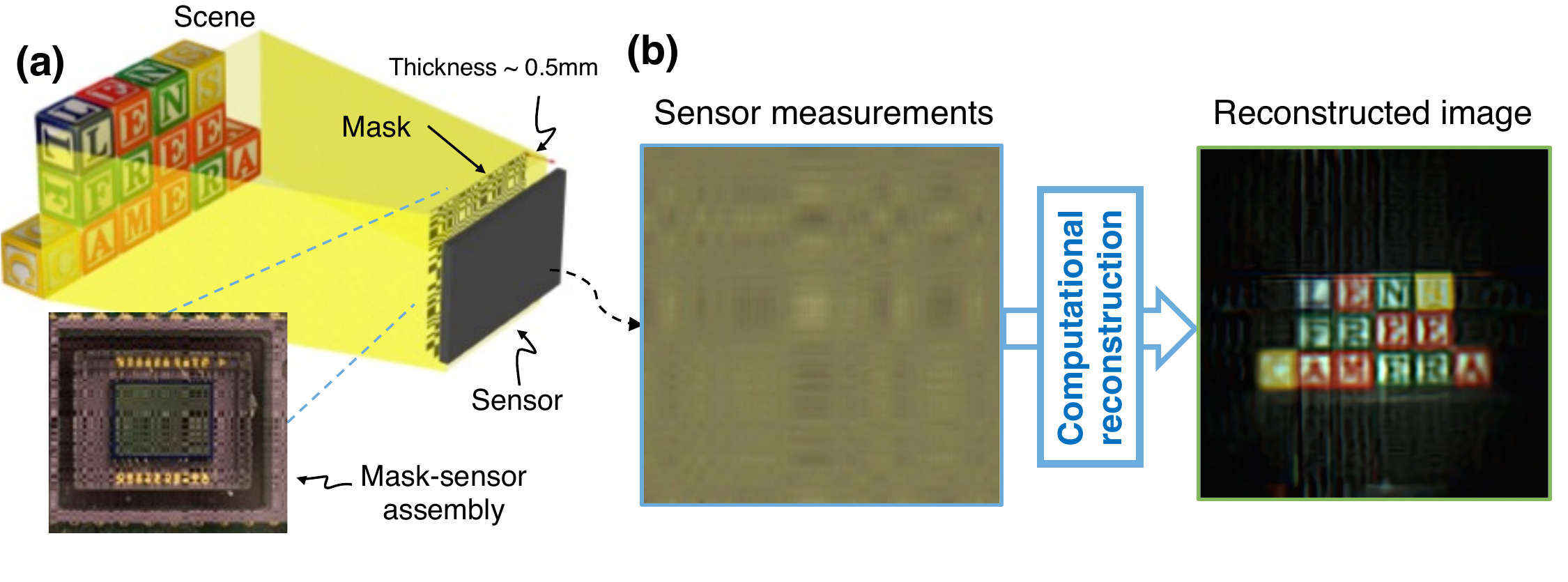}
	\caption{{FlatCam architecture.} \textbf{(a)} Every light source within the camera field-of-view contributes to every pixel in the multiplexed image formed on the sensor. A computational algorithm reconstructs the image of the scene. Inset shows the mask-sensor assembly of our prototype in which a binary, coded mask is placed 0.5mm away from an off-the-shelf digital image sensor. \textbf{(b)} An example of sensor measurements and the image reconstructed by solving a computational inverse problem.}
	\label{fig:flatcam}
\end{figure*}

An illustration of the FlatCam design is presented in Fig.~\ref{fig:flatcam}. Light from a scene passes through a coded mask and lands on a conventional image sensor. The mask consists of opaque and transparent features (to block and transmit light, respectively); each transparent feature can be viewed as a pinhole. Light from the scene gets diffracted by the mask features such that light from each scene location casts a unique mask shadow on the sensor, and this mapping can be represented using a linear operator. A computational algorithm then inverts this linear operator to recover the original light distribution of the scene from the sensor measurements. 

Our FlatCam design has many attractive properties besides its slim profile. First, since it reduces the thickness of the camera but not the area of the sensor, it collects more light than miniature, lens-based cameras with same thickness. Second, the mask can be created from inexpensive materials that operate over a broad range of wavelengths. Third, the mask can be fabricated simultaneously with the sensor array, creating new manufacturing efficiencies. The mask can be fabricated either directly in one of the metal interconnect layers on top of the photosensitive layer or on a separate wafer thermal compression that is bonded to the back side of the sensor, as is typical for back-side illuminated image sensors \cite{dragoi2010cmos}. 

We demonstrate the potential of the FlatCam using two prototypes built in our laboratory with commercially available sensors and masks: a visible prototype in which the mask-sensor spacing is about 0.5mm and a short-wave infrared (SWIR) prototype in which the spacing is about 5mm. Figures \ref{fig:visible} and \ref{fig:swir} illustrate sensor measurements and reconstructed images using our prototype FlatCams.

\section{Related work}
\textit{Pinhole cameras.} Imaging without a lens is not a new idea. Pinhole cameras, the progenitor of lens-based cameras, have been well known since Alhazen (965--1039AD) and Mozi (c. 370BCE).  However, a tiny pinhole drastically reduces the amount of light reaching the sensor, resulting in noisy, low-quality images. Indeed, lenses were introduced into cameras for precisely the purpose of increasing the size of the aperture, and thus the light throughput, without degrading the sharpness of the acquired image. 

\begin{figure*}[ttt]
	\centering
	\includegraphics[width=0.95\textwidth,trim = 0mm 70mm 0mm 0mm, clip, page=2]{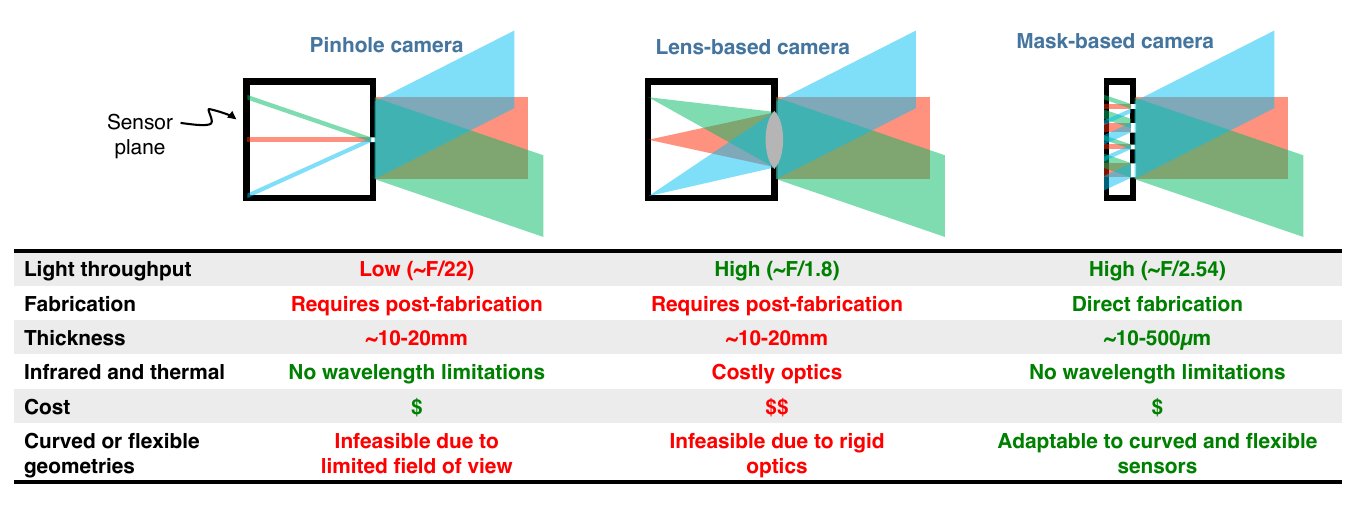}
	\caption{Comparison of pinhole, lens-based, and coded mask-based cameras. Pinhole cameras and lens-based cameras provide one-to-one mapping between light from a focal plane and the sensor plane (note that light from three different directions is mapped to three distinct locations on the sensor), but the coded mask-based cameras provide a multiplexed image that must be resolved using computation. The table highlights some salient properties of the three camera designs. Pinholes cameras suffer from very low light throughput, while lens-based cameras are bulky and rigid because of their optics. In contrast, the FlatCam design offers thin, light-efficient cameras with the potential for direct fabrication.}
	\label{fig:comparison}
\end{figure*}

\textit{Coded apertures.}  Coded aperture cameras extend the idea of a pinhole camera by using masks with multiple pinholes  \cite{dicke1968scatter,fenimore1978coded,cannon1980coded}.
The primary goal of coded aperture cameras is to increase the light throughput compared to a  pinhole camera. 
Figure~\ref{fig:comparison} summarizes some salient features of pinhole, lens-based, and FlatCam (coded mask-based) architectures.

Coded-aperture cameras have traditionally been used for imaging wavelengths beyond the visible spectrum (e.g., x-ray and gamma-ray imaging), for which lenses or mirrors are expensive or infeasible \cite{dicke1968scatter,fenimore1978coded,cannon1980coded,durrant1999application,brady2009optical}. 
Mask-based lens-free designs have also been proposed for flexible field-of-view selection in \cite{zomet2006lensless}, compressive single-pixel imaging using a transmissive LCD panel \cite{huang2013lensless}, and separable coded masks \cite{deweert2015lensless}. 

In recent years, coded masks and light modulators have also been added to lens-based cameras in different configurations to build novel imaging devices that can capture image and depth \cite{levin2007image}, dynamic video \cite{liu_PCeV2013}, or 4D lightfield \cite{veeraraghavan2007dappled,marwah2013compressive} from a single coded image. 
Coded aperture-based systems using compressive sensing principles \cite{candes2006stable,donoho2006compressed,baraniuk2007compressive} have also been studied for image super-resolution \cite{marcia2008compressive}, spectral imaging \cite{wagadarikar2008single}, and video capture \cite{llull2013coded}. 

Existing coded aperture-based lensless systems have two main limitations: 
First, the large body of work devoted to coded apertures invariably place the mask significantly far away from the sensor (e.g., 65mm distance in \cite{deweert2015lensless}). 
In contrast, our FlatCam design offers a thin form factor. For instance, in our prototype with a visible sensor, the spacing between the sensor and the mask is only 0.5mm.  
Second, the masks employed in some designs have transparent features only in a small central region whose area is invariably much smaller than the area of the sensor. 
In contrast, almost half of the features (spread across the entire surface) in our mask are transparent.
As a consequence, the light throughput of our designs are many orders of magnitude larger as compared to previous designs.
Furthermore, the lensless cameras proposed in  \cite{huang2013lensless,deweert2015lensless} use programmable spatial light modulators (SLM) and capture multiple images while changing the mask patterns. 
In contrast, we use a static mask in our design, which can potentially be fixed on the sensor during fabrication or the assembly process. 

\textit{Camera arrays.} A number of thin imaging systems have been developed over the last few decades. The TOMBO architecture \cite{tanida2001thin}, inspired by insect compound eyes, reduces the camera thickness by replacing a single, large focal-length lens with multiple, small focal-length microlenses. Each microlens and the sensor area underneath it can be viewed as a separate low-resolution, lens-based camera, and a single high-resolution image can be computationally reconstructed by fusing all of the sensor measurements. Similar architectures have been used for designing thin infrared cameras \cite{shankar2008thin}. The camera thickness in this design is dictated by the geometry of the microlenses; reducing the camera thickness requires a proportional reduction in the sizes of the microlenses and sensor pixels. As a result, microlens-based cameras currently offer only up to a four-fold reduction in the camera thickness \cite{bruckner2010thin,venkataraman2013picam}. 

\textit{Folded optics.} An alternate approach for achieving thin form factors relies on folded optics, where light manipulation similar to that of a traditional lens is achieved using multi-fold reflective optics \cite{tremblay2007ultrathin}. However, folded optics based systems have low light collection efficiencies. 

\textit{Ultra-miniature lensless imaging with diffraction gratings.} Recently, miniature cameras with integrated diffraction gratings and CMOS image sensors have been developed \cite{wang2009angle,gill2011microscale,gill2013lensless,stork2013sensorcomm}. These cameras have been successfully demonstrated on tasks such as motion estimation and face detection. While these cameras are indeed ultra-miniature in total volume (100 micron sensor width by 200 micron thickness), they retain the large thickness-to-width ratio of conventional lens-based cameras. Because of the small sensor size, they suffer from reduced light collection ability. In contrast, in our visible prototype below, we used a 6.7mm wide square sensor, which increases the amount of light collection by about three orders of magnitude, while the device thickness remains approximately similar (500 micron).

\textit{Lensfree microscopy and shadow imaging.} Lensfree cameras have been successfully demonstrated for several microscopy and lab-on chip application, wherein the subject to be imaged is close to the image sensor.
An on-chip, lens-free microscopy design that uses amplitude masks to cast a shadow of point illumination sources onto a microscopic tissue sample has shown significant promise for microscopy and related applications, where the sample being imaged is very close to the sensor (less than 1mm) \cite{greenbaum2012imaging,greenbaum2014wide}. Unfortunately, this technique cannot be directly extended to traditional photography and other applications that require larger standoff distances and do not provide control over illumination.

\section{FlatCam design}
Our FlatCam design places an amplitude mask almost immediately in front of the sensor array (see Fig.~\ref{fig:flatcam}). We assume that the sensor and the mask are planar, parallel to each other, and separated by distance  $d$. 
While we focus on a single mask for exposition purposes, the concept extends to multiple amplitude masks in a straightforward manner.  
For simplicity of explanation, we also assume (without loss of generality) that the mask modulates the impinging light in a binary fashion; that is, it consists of transparent features that transmit light and opaque features that block light. We denote the size of the transparent/opaque features by $\Delta$ and assume that the mask covers the entire sensor array. 

\add{
	Consider the one-dimensional coded aperture system depicted in Fig.~\ref{fig:CA_cra}, in which a single coded mask is placed at distance $d$ from the sensor plane. We assume that the field-of-view (FOV) of each sensor pixel is limited by a chief ray angle (CRA) $\theta_\text{CRA}$, which implies that every pixel receives light only from the angular directions that lie within $\pm \theta_\text{CRA}$ with respect to its surface normal. Therefore, light rays entering any pixel are modulated by the mask pattern of length $w=2d\tan \theta_\text{CRA}$. As we increase (or decrease) the mask-to-sensor distance, $d$, the width of the mask pattern, $w$, also increases (or decreases). Assuming that the scene is far from the camera, the mask patterns for neighboring pixels shift by the same amount as the pixel width. Therefore, if we assume that the mask features and the sensor pixels have the same width, $\Delta$, then the mask patterns for neighboring pixels shift by exactly one mask element. Furthermore, if we fix $d\approx N\Delta/2\tan\theta_\text{CRA}$, then exactly $N$ mask features lie within the FOV of each pixel. If the mask is designed by repeating a pattern of $N$ features, then the linear system that maps the light distribution in the scene to the sensor measurements can be represented as a cyclic convolution.
	
	\begin{figure}[t]
		\centering
		\includegraphics[width=0.75\linewidth, page=3, trim=0mm 115mm 100mm 0mm, clip]{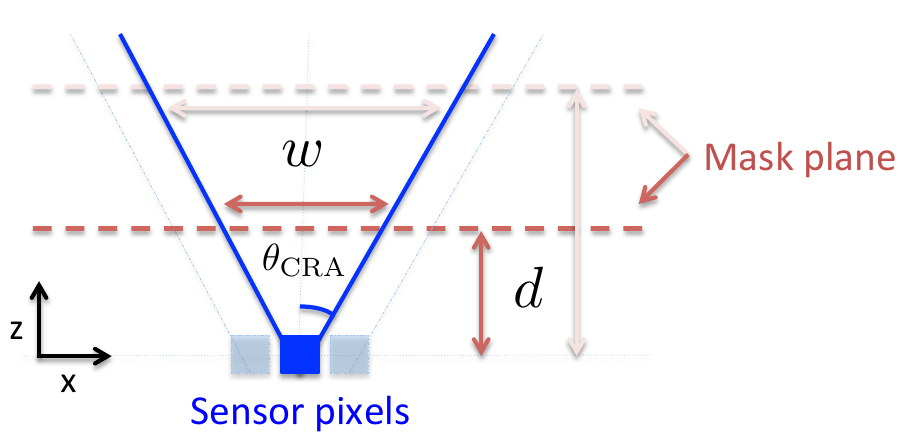}
		\caption{An illustration of a coded aperture system with a mask placed $d$ units away from the sensor plane. Each pixel records light from angular directions within $\pm\theta_\text{CRA}$. Light reaching each sensor is modulated by the mask pattern that is $w=2d\tan \theta_\text{CRA}$ units wide, which we can increase (or decrease) by moving the mask farther (or closer) to the sensor. }
		\label{fig:CA_cra}
	\end{figure}
	
	A number of mask patterns have been introduced in the literature that offer high light collection and robust image reconstruction for circulant systems. Typical examples  include uniform redundant array (URA), modified URA (MURA), and pseudo-noise pattens such as maximum length sequences (MLS or M-sequences) \cite{macwilliams1976pseudo,fenimore1978coded,gottesman1989new,busboom1998uniformly,golomb1982shift}. One key property of these patterns is that they have near-flat Fourier spectrum. Furthermore, coded aperture systems have been conventionally used for imaging X rays and Gamma rays for which diffraction effects can be ignored. Therefore, if we can align the mask according to the ideal configuration and ignore the diffraction effects, then the system response is a cyclic convolution.
	
	Our FlatCam design does not necessarily yield a circulant system because our goal is to use the smallest possible mask-to-sensor distance, $d$. Furthermore, to build our prototypes, we used the smallest mask feature size, $\Delta$, for which we could manually align the system on the optical table using mechanical posts. However, we demonstrate that by employing a scalable calibration procedure with a separable mask pattern in our FlatCam design, we can reconstruct quality images via simple computational algorithms. 
}

\subsection{Replacing lenses with computation} 
Light from all points in the three dimensional scene gets modulated and diffracted by the mask pattern and recorded on the image sensor. 
\add{
	Let us consider a surface, $\mathcal{S}$, in the scene that is completely visible to the sensor pixels and denote $x$ as a vector of light distribution from all the points in $\mathcal{S}$. We can then describe the sensor measurements, $y$, as 
	\begin{equation}\label{eq:y=Phi x+e}
	y = \Phi x + e.  
	\end{equation}
	$\Phi$ denotes a transfer matrix whose $i$th column corresponds to an image that would form on the sensor if the scene contains a single light source of unit intensity at $i$th location in $x$. $e$ denotes the sensor noise and any model mismatch. 
	Note that if all the points in the scene, $x$, are at the same depth, then $\mathcal{S}$ becomes a plane parallel to the mask at distance $d$. 
}
Since the sensor pixels do not have a one-to-one mapping with the scene pixels, the matrix $\Phi$ will not resemble the identity matrix. Instead, each sensor pixel measures multiplexed light from multiple scene pixels, and each row of $\Phi$ indicates how strongly each scene pixel contributes to the intensity measured at a particular sensor pixel. In other words, any column in $\Phi$ denotes the image formed on the sensor if the scene contains a single, point light source at the respective location. 


Multiplexing generally results in an ill-conditioned system. Our goal is to design a mask that produces a matrix $\Phi$ that is well conditioned and hence can be stably inverted without excessive noise amplification. We now discuss how we navigate among three inter-related design decisions: the mask pattern, the placement $d$ and feature size $\Delta$ of the mask, and the image recovery (demultiplexing) algorithm.

\subsection{Mask pattern} 
\add{
	The design of mask patterns plays an important role in coded-aperture imaging. An ideal pattern would maximize the light throughput while providing a well-conditioned scene-to-sensor transfer functions. In this regard, notable examples of mask patterns include URA, MURA, and pseudo noise patterns  \cite{macwilliams1976pseudo,fenimore1978coded,gottesman1989new}. URAs are particularly useful because of two key properties: (1) almost half of the mask is open, which helps with the signal-to-noise ratio; (2) the autocorrelation function of the mask is close to a delta function, which helps in image reconstruction. URA patterns are closely related to the Hadamard-Walsh pattern and the MLS that are maximally incoherent with their own cyclic shifts \cite{fenimore1981fast,busboom1998uniformly,ding2015complementary}. However, these properties only hold true if the mask and sensor can be aligned to create a perfect circulant system. 
}

In FlatCam design we consider three parameters to select the mask pattern: the light throughput, the complexity of system calibration and inversion, and the conditioning of the resulting multiplexing matrix $\Phi$.

\textit{Light throughput.}
In the absence of the mask, the amount of light that can be sensed by the bare sensor is limited only by its CRA. Since the photosensitive element in a CMOS/CCD sensor array is situated in a small cavity, a micro-lens array directly on top of the sensor is used to increase the light collection efficiency. In spite of this, only light rays up to a certain angle of incidence reach the sensor, and this is the fundamental light collection limit of that sensor. Placing an amplitude-modulating mask very close to (and completely covering) the sensor results in a light-collection efficiency that is a fraction of the fundamental light collection limit of the sensor. In our designs, half of the binary mask features are transparent, which halves our light collection ability compared to the maximum limit. 

\begin{figure*}
	\centering
	\begin{subfigure}[t]{0.3\textwidth}
		\includegraphics[width=1\textwidth, page=6, trim=0mm 20mm 0mm 0mm, clip]{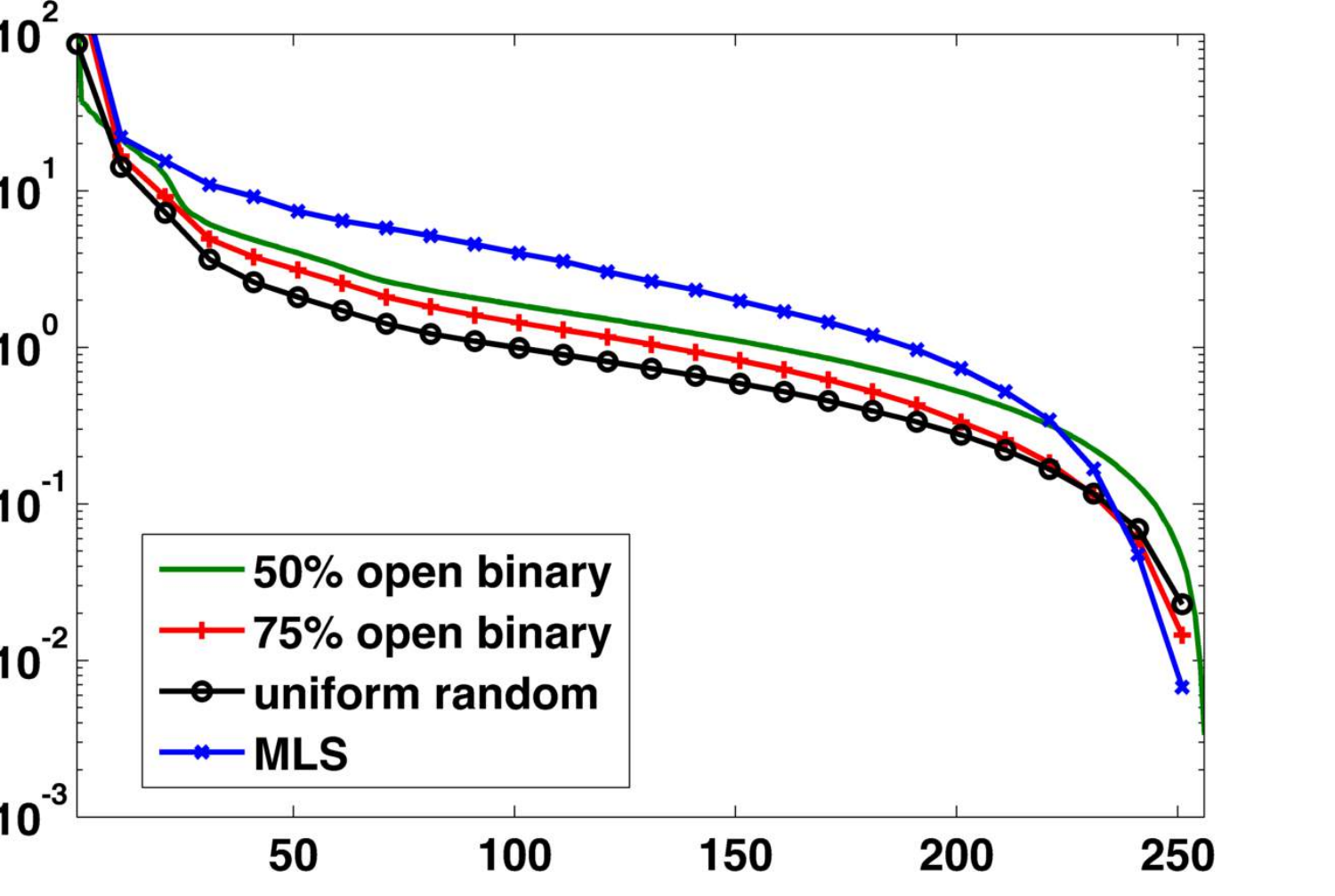}
		\caption{Effect of different transparent features  }\label{fig:transparencyLevel}
	\end{subfigure}
	\begin{subfigure}[t]{0.3\textwidth}
		\includegraphics[width=1\textwidth, page=4, trim=0mm 20mm 0mm 0mm, clip]{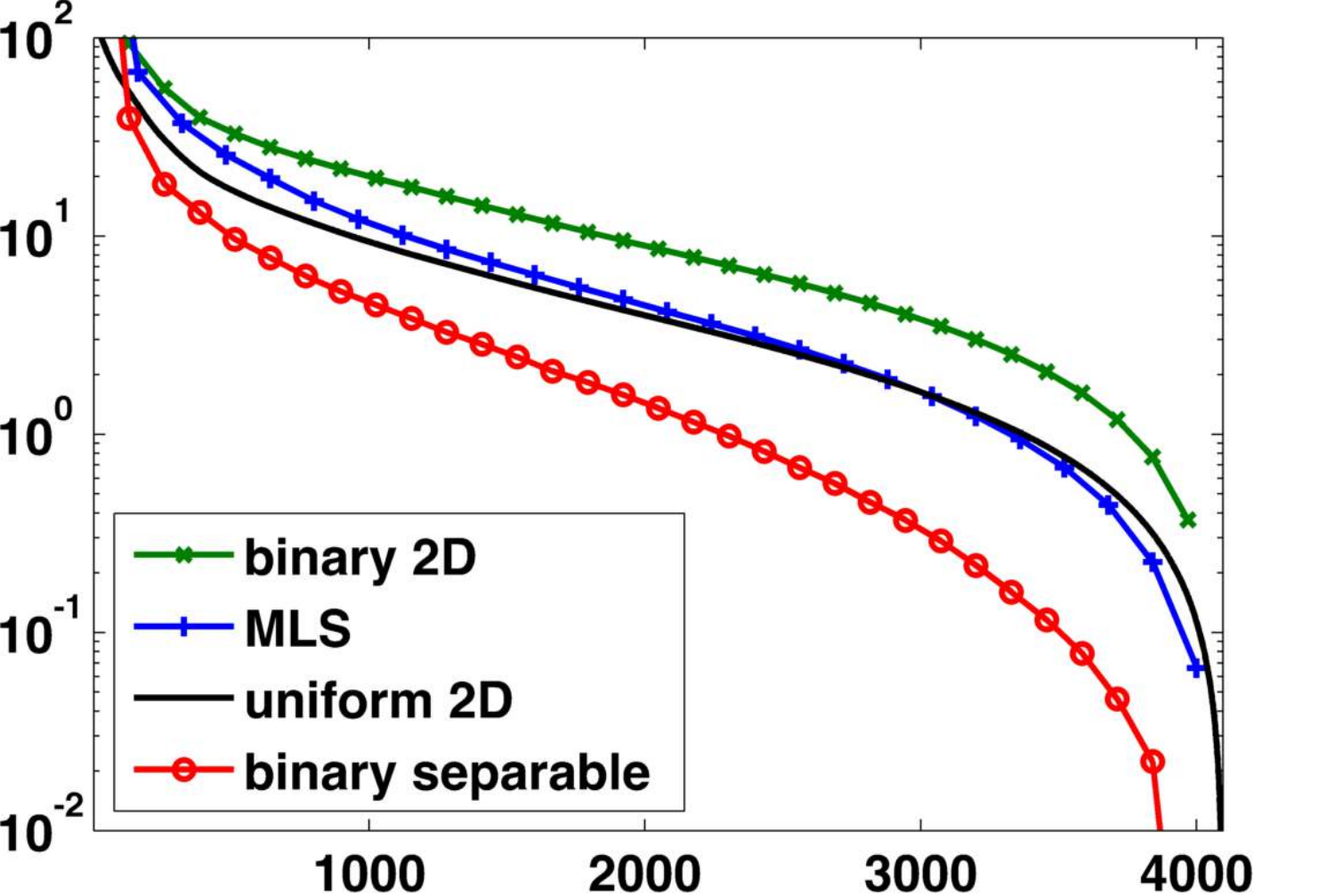}
		\caption{Separable vs non-separable patterns} \label{fig:sepVsnonsep}
	\end{subfigure}
	\begin{subfigure}[t]{0.3\textwidth}
		\includegraphics[width=1\textwidth, page=10, trim=0mm 20mm 0mm 0mm, clip]{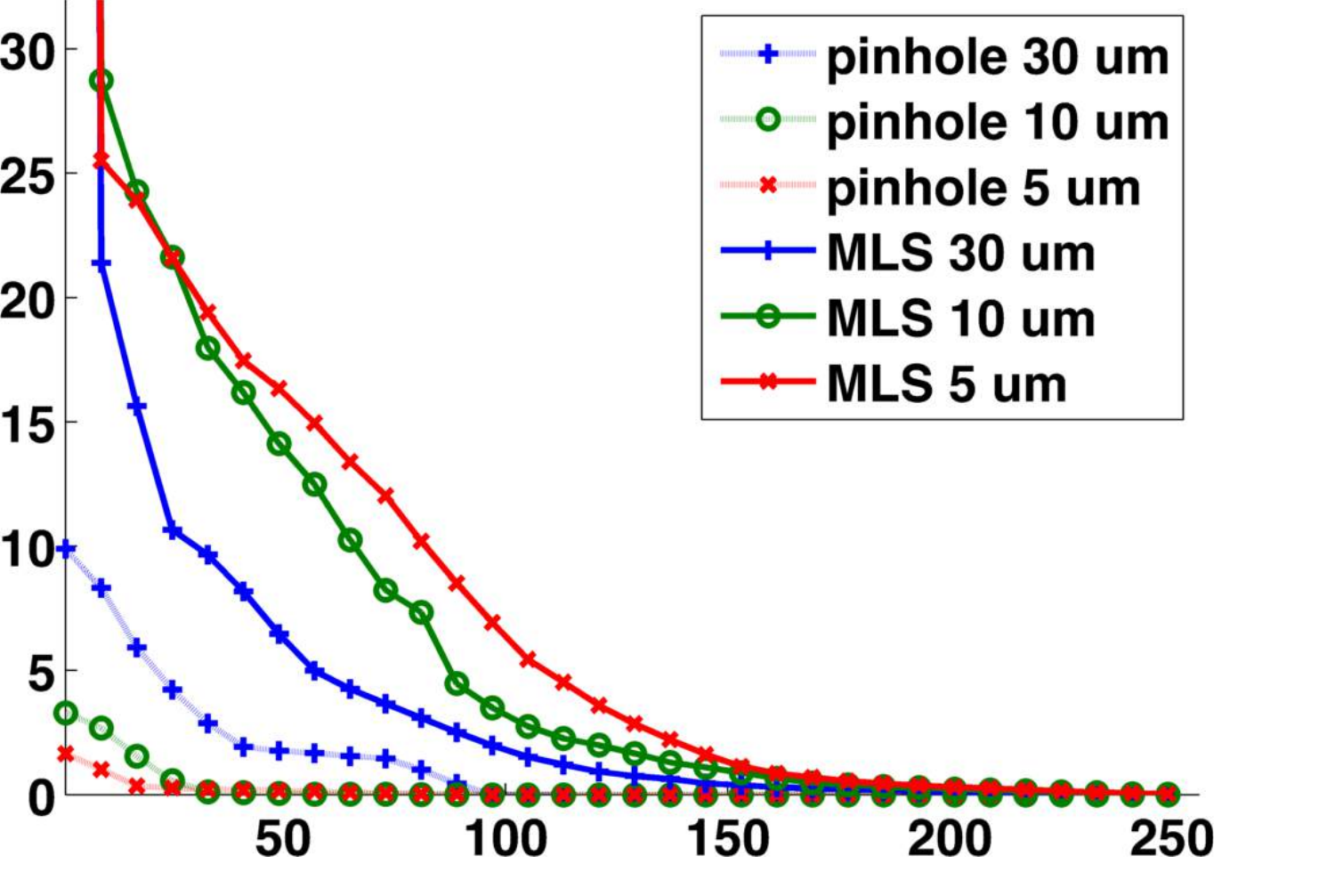}
		\caption{Pinhole vs coded masks} \label{fig:mlsVspinholes}
	\end{subfigure}
	\vspace{10pt}
	\caption{Analysis of coded aperture system singular values by simulating sensing matrix with different masks placed at $d=500\,\mu$m. 
		\textbf{(a)} Increasing the number of transparent features beyond $50\%$ may increase light collection, but it deteriorates the conditioning of the system. 
		\textbf{(b)} Even though a non-separable pattern may provide better reconstruction compared to a separable pattern, calibrating and characterizing such a system requires a highly sophisticated calibration procedure.
		\textbf{(c)} A single pinhole-based system degrades as we reduce the feature size because lesser light reaches the sensor and the blur size increases. In contrast, a coded mask-based system improves as we reduce the feature size. 
	}
\end{figure*}

\add{ 
	To compare mask patterns with different types of transparent features, we present a simulation result in Fig.~\ref{fig:transparencyLevel}. We simulated the transfer matrix, $\Phi$, for a one-dimensional system for four different types of masks and compared the singular values of their respective $\Phi$. Ideally, we want a mask for which the singular values of $\Phi$ are large and they decay at a slow rate. We generated one-dimensional mask patterns using random binary patterns with $50\%$ and $75\%$ open features, uniform random pattern with entries drawn from the unit interval, $[0,1]$, and an MLS pattern with $50\%$ open features. We observed that MLS pattern outperforms random patterns and increasing the number of transparent features beyond $50\%$ deteriorates the conditioning of the system. 
}

As described above, while it is true that the light collection ability of our FlatCam design is one-half of the maximum achievable with a particular sensor, the main advantage of the FlatCam design is that it allows us to use much larger sensor arrays for a given device thickness constraint, thereby significantly increasing the light collection capabilities of devices under thickness constraints.

\textit{Computational complexity.} 
The (linear) relationship between the scene irradiance $x$ and the sensor measurements $y$  is contained in the multiplexing matrix $\Phi$. Discretizing the unknown scene irradiance into $N\times N$ pixel units and assuming an $M\times M$ sensor array, $\Phi$ is an $M^2\times N^2$ matrix. Given a mask and sensor, we can obtain the entries of $\Phi$ either by modeling the transmission of light from the scene to the sensor or through a calibration process.  Clearly, even for moderately sized systems, $\Phi$ is prohibitively large to either estimate (calibration) or invert (image reconstruction), in general. For example, to describe a system with a megapixel resolution scene and a megapixel sensor array, $\Phi$ will contain on the order of $10^6\times 10^6=10^{12}$  elements. 

One way to reduce the complexity of $\Phi$ is to use a separable mask for the FlatCam system. If the mask pattern is separable (i.e., an outer product of two one-dimensional patterns), then the imaging system in \eqref{eq:y=Phi x+e} can be rewritten as 
\begin{equation}\label{eq:Y=PL X PR}
Y = \Phi_L X \Phi_R^T + E,
\end{equation}
where $\Phi_L,\Phi_R$ denote matrices that correspond to one-dimensional convolution along the rows and columns of the scene, respectively, $X$ is an $N\times N$ matrix containing the scene radiance, $Y$ in an $M\times M$ matrix containing the sensor measurements, and $E$ denotes the sensor noise and any model mismatch. 
For a megapixel scene and a megapixel sensor, $\Phi_L$ and $\Phi_R$ have only $10^6$ elements each, as opposed to $10^{12}$ elements in $\Phi$.
Similar idea has been recently proposed in \cite{deweert2015lensless} with the design of doubly Toeplitz mask. 
In our implementation, we also estimate the system matrices using a separate calibration procedure (see Sec.~\ref{sec:calibration}), which also becomes significantly simpler for a separable system.  

\add{ 
	To compare separable and non-separable mask patterns, we present a simulation result in Fig.~\ref{fig:sepVsnonsep}. We simulated the $\Phi$ matrices for a two-dimensional scene at $64\times 64$ resolution using two separable and two non-separable mask patterns and compared the singular values of their respective $\Phi$.
	For non-separable mask patters, we generated a random binary 2D pattern with  an equal number of 0,1 entries and a uniform 2D pattern with entries drawn uniformly from the unit interval. 
	For separable mask patterns, we generated an MLS pattern by first computing an outer product of two one-dimensional M-sequences with $\pm1$ entries and setting all $-1$s to zero, and a separable binary pattern by computing the outer product of two one-dimensional binary patterns so that the number of 0s and 1s in the resulting 2D pattern is almost the same.
	Even though the non-separable binary pattern provides better singular values compared to a separable MLS pattern, calibrating and characterizing such a system for high-dimensional images is beyond our current capabilities.
}

\textit{Conditioning.} 
The mask pattern should be chosen to make the multiplexing matrices $\Phi_L$ and $\Phi_R$ as numerically stable as possible, which ensures a stable recovery of the image $X$ from the sensor measurements $Y$. Such $\Phi_L$ and $\Phi_R$ should have low condition numbers, i.e., a flat singular value spectrum. For Toeplitz matrices, it is well known that, of all binary sequences, the so-called maximum length sequences, or M-sequences, have maximally flat spectral properties \cite{golomb1982shift}. Therefore, we use a separable mask pattern that is the outer product of two one-dimensional M-sequence patterns. However, because of the inevitable non-idealities in our implementation, such as the limited sensor CRA and the larger than optimal sensor-mask distance due to the hot mirror, the actual $\Phi_L$ and $\Phi_R$ we obtain using a separable M-sequence based mask do not achieve a perfectly flat spectral profile. Nevertheless, as we demonstrate in our prototypes, the resulting multiplexing matrices enable stable image reconstruction in the presence of sensor noise and other non-idealities. All of the visible wavelength, color image results shown in this paper were obtained using normal, indoor ambient lighting and exposure times in 10--20ms range, demonstrating that robust reconstruction is possible.
 
\subsection{Mask placement and feature size} 
The multiplexing matrices $\Phi_L,\Phi_R$ describe the mapping of light emanating from the points in the scene to the pixels on the sensor. Consider light from a point source passing through one of the mask openings; its intensity distribution recorded at the sensor forms a point-spread function (PSF) that is due to both diffraction and geometric blurs. The PSF acts as a low-pass filter that limits the frequency content that can be recovered from the sensor measurements. The choice of the feature size and mask placement is dictated by the tradeoff between two factors: reducing the size of the PSF to minimize the total blur and enabling sufficient multiplexing to obtain a well-conditioned linear system. 

The total size of the PSF depends on the diffraction and geometric blurs, which in turn depend on the distance between the sensor and the mask, $d$, and the mask feature size, $\Delta$. The size of the diffraction blur is approximately $2.44 \lambda d/\Delta$, where $\lambda$ is the wavelength of light waves. The size of the geometric blur, however, is equal to the feature size $\Delta$. Thus, the minimum blur radius for a fixed $d$ is achieved when the two blur sizes are approximately equal: $\Delta = \sqrt{2.44 \lambda d}$. One possible way to reduce the size of the combined PSF is to use larger feature size $\Delta$. However, the extent of multiplexing within the scene pixels reduces as $\Delta$ increases. Therefore, if we aim to keep the amount of multiplexing constant, then the mask feature size $\Delta$ should shrink proportionally to the mask-sensor distance $d$.
 
In practice, physical limits on the sensor-mask distance $d$ or the mask feature size $\Delta$ can dictate the design choices. In our visible FlatCam prototype, for example, we use a Sony ICX285 sensor. The sensor has a 0.5 mm thick hot mirror attached to the top of the sensor, which restricts the potential spacing between the mask and sensor surface. Therefore, we attach the mask to the hot mirror, resulting in $d \approx 500\,\mu$m (distance between the mask and the sensor surface). For a single pinhole at this distance, we achieve the smallest total blur size using a mask feature size of approximately $\Delta = 30\,\mu$m, which is also the smallest feature size for which we were able to properly align the mask and sensor on the optical table. 
Of course, in future implementations, where the mask pattern is directly etched on top of the image sensor (direct fabrication) such practical constraints can be alleviated and we can achieve much higher resolution images by moving the mask closer to the sensor and reducing the mask feature size proportionally.

\add{ 
	To compare the effect of feature size on the conditioning of the sensing matrix, $\Phi$, we present a simulation result in Fig.~\ref{fig:mlsVspinholes}. We simulated the $\Phi$ matrices for an MLS mask and a single pinhole for three different values of $\Delta = \{5, 10, 30\}\,\mu$m. 
	For a pinhole pattern, we observe that reducing the pinhole size, $\Delta$, degrades conditioning of $\Phi$ in two ways: (1) The largest singular value of $\Phi$ reduces as lesser light passes through the pinhole. (2) The singular values decay faster as the total blur increases because of smaller pinholes. 
	For an MLS pattern, we observed that reducing the feature size, $\Delta$, improves the conditioning of the system matrix $\Phi$. However,  because of the practical challenges we encountered while aligning mask and sensor, we do not yet have an experimental evidence to support the simulation results. 
}

\subsection{Camera calibration}\label{sec:calibration}
We now provide the details of our calibration procedure for the separable imaging system modeled in  \eqref{eq:Y=PL X PR}. Instead of modeling the convolution shifts and diffraction effects for a particular mask-sensor arrangement, we directly estimate the system matrices. 

To align the mask and sensor, we adjust their relative orientation such that a separable scene in front of the camera yields a separable image on the sensor. 
For a coarse alignment, we use a point light source, which projects a shadow of the mask onto the sensor, and align the horizontal and vertical edges on the sensor image with the image axes. 
For a fine alignment, we align the sensor with the mask while projecting horizontal and vertical stripes on a monitor or screen in the front of the camera.

\begin{figure}[t]
	\centering
	\includegraphics[width=0.95\linewidth,trim = 0mm 10mm 0mm 0mm, clip, page=5]{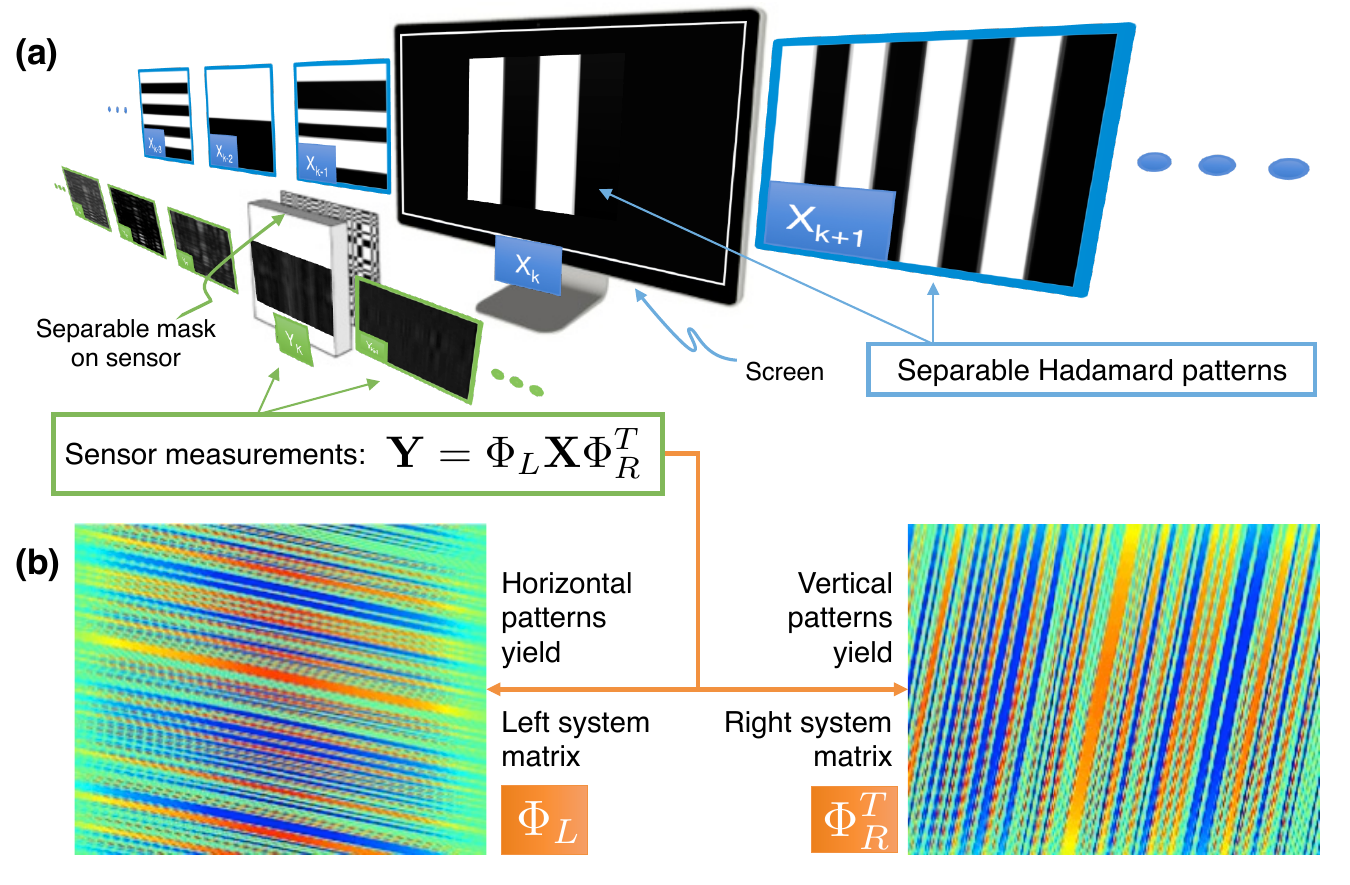}
	\caption{Calibration for measuring the left and right multiplexing matrices $\Phi_L$ and $\Phi_R$ corresponding to a separable mask. \textbf{(a)} Display separable patterns on a screen. The patterns are orthogonal, one-dimensional Hadamard codes that are repeated along either the horizontal or vertical direction. \textbf{(b)} Estimated left and right system matrices.}
	\label{fig:calibration}
\end{figure}

To calibrate a system that can recover $N\times N$ images $X$, we estimate the left and right matrices $\Phi_L,\Phi_R$ using the sensor measurements of $2N$ known calibration patterns projected on a screen as depicted in Fig.~\ref{fig:calibration}. 
Our calibration procedure relies on an important observation.
If the scene $X$ is separable, i.e., $X = {\bf a b}^T$ where ${\bf a}, {\bf b} \in \mathbb{R}^N$, then
\[ 
Y = \Phi_L {\bf a b}^T \Phi_R^T  = (\Phi_L {\bf a}) (\Phi_R {\bf b})^T. 
\]
In essence, the image formed on the sensor is a rank-1 matrix, and by using a truncated singular value decomposition (SVD), we can obtain $\Phi_L {\bf a}$ and $\Phi_R {\bf b}$ up to a signed, scalar constant.
We take $N$ separable pattern measurements for calibrating each of $\Phi_L$ and $\Phi_R$.

Specifically, to calibrate $\Phi_L$, we capture $N$ images $\{Y_1, \ldots, Y_N \}$ corresponding to the separable patterns $\{X_1, \ldots, X_N\}$ displayed on a monitor or screen.
Each $X_k$ is of the form
$ X_k = {\bf h}_k {\bf 1}^T,$
where ${\bf h}_k \in \mathbb{R}^N$ is a column of the orthogonal Hadamard matrix $H$ of size $N \times N$ and ${\bf 1}$ is an all-ones vector of length $N$. 
Since the Hadamard matrix consists of $\pm1$ entries, we record two images for each Hadamard pattern; one with ${\bf h}_k{\bf1}^T$ and one with $-{\bf h}_k{\bf1}^T$ while setting the negative entries to zero in both cases.
We then subtract the two sensor images to obtain the measurements corresponding to $X_k$. 
Let  $\widetilde{Y}_k = {\bf u}_k {\bf v}^T$ be the rank-1 approximation of the measurements $Y_k$ obtained via SVD, where the underlying assumption is that ${\bf v}\approx\Phi_R{\bf 1}$, upto a signed, scalar constant.
Then, we have
\begin{equation}
[{\bf u}_1\,{\bf u}_2\, \cdots{\bf u}_N] = \Phi_L  [ {\bf h}_1\,{\bf h}_2\, \cdots{\bf h}_N]\equiv \Phi_L H, \\
\end{equation}
and we compute $\Phi_L$ as 
\begin{equation}\label{eq:calibComputation}
\Phi_L = [{\bf u}_1\,{\bf u}_2\, \cdots{\bf u}_N] H^{-1}, 
\end{equation}
where $H^{-1}= \frac{1}{N}H^T$. 
Similarly, we estimate $\Phi_R$ by projecting $N$ patterns of the form ${\bf 1} {\bf h}_k^T$.

Figure~\ref{fig:calibration} depicts the calibration procedure in which we projected separable patterns on a screen and recorded sensor measurements; the sensor measurements recorded from these patterns are re-ordered to form the left and right multiplexing operators shown in {(b)}.

A mask can only modulate light with non-negative transmittance values. 
M-sequences are defined in terms of $\pm 1$ values and hence cannot be directly implemented in a mask.
The masks we use in our prototype cameras are constructed by computing the outer-product of two M-sequences and then setting the resulting $-1$ entries to $0$.
%
%
This produces a mask that is optically feasible but no longer mathematically separable.
We can resolve this issue in post-processing, since the difference between the measurements using the theoretical $\pm1$ separable mask and the optically feasible $0/1$ mask is simply a constant bias term.
In practice, once we acquire a sensor image, we correct it to correspond to a $\pm 1$ separable mask (described as $Y$ in \eqref{eq:Y=PL X PR}) simply by forcing the column and row sums to zero. 

\begin{figure}[t]
	\centering
	\includegraphics[width=1\linewidth, page=1, trim=0mm 55mm 0mm 0mm, clip]{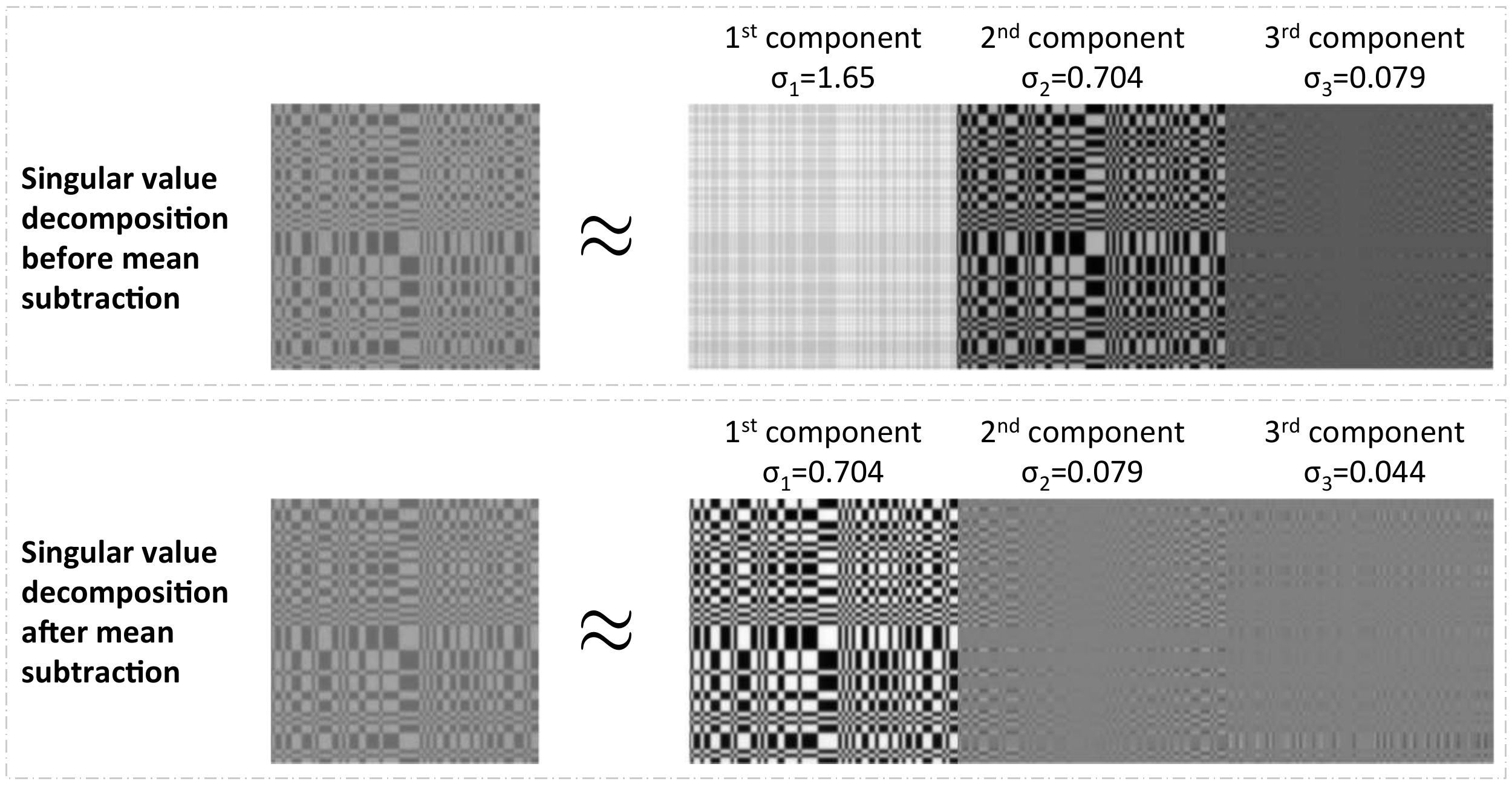}
	\caption{Singular value decomposition of the point spread function for our proposed separable MLS mask before and after mean subtraction.}
	\label{fig:separableMean}
\end{figure}

\add{
	Recall that if we use a separable mask, then we can describe sensor measurements as $Y = \Phi_L X \Phi_R^T$. If we turn on a single pixel in $X$, say $X_{ij} = 1$, then the sensor measurements would be a rank-1 matrix $\phi_i\phi_j^T$, where $\phi_i, \phi_j$ denote the $i$th and $j$th columns in $\Phi_L,\Phi_R$, respectively. 
	Let us denote $\psi$ as a one-dimensional M-sequence of length $N$ and $\Psi_{\pm1}=\psi\psi^T$ as the separable mask that consists of $\pm1$ entries; it is optically infeasible because we cannot subtract light intensities. We created an optically feasible mask by setting all the $-1$s in $\Psi_{\pm1}$ to $0$s, which can be described as $$\Psi_{0/1} = (\Psi_{\pm 1} + \mathbf{11}^T)/2.$$ 
	Therefore, if we have a single point source in the scene, the sensor image  will be a rank-2 matrix. By subtracting the row and column means of the sensor image, we can convert the sensor response back to a rank-1 matrix. Only after this correction can we represent the superposition of sensor measurements from all the light sources in the scene using the separable system $Y=\Phi_L X \Phi_R^T$.  We present an example of this mean subtraction procedure for an image captured with our prototype and a point light source in Fig.~\ref{fig:separableMean}. 
}

\section{Image reconstruction}
Given a set of $M\times M$ sensor measurements $Y$, our ability to invert the system \eqref{eq:Y=PL X PR} to recover the desired $N\times N$ image $X$ primarily depends on the rank and the condition number of the system matrices $\Phi_L$, $\Phi_R$.

If both $\Phi_L$ and $\Phi_R$ are well-conditioned, then we can estimate $X$ by solving a simple least-squares problem
\begin{equation}\label{eq:LS}
\widehat{X}_{\textrm{LS}} = \arg \min_X\; \|\Phi_L X\Phi_R^T -Y\|_2^2,
\end{equation}
which has the closed form solution: $\widehat{X}_{\textrm{LS}} = \Phi_L^{+} Y\Phi_R^{+}$, where $\Phi^+_L$ and $\Phi_R^+$ denote the pseudoinverse of $\Phi_L$ and $\Phi_R$, respectively. 
Consider the SVD of $\Phi_L = U_L\Sigma_L V_L^T$, where $U_L$ and $V_L$ are orthogonal matrices that contain the left and right singular vectors and $\Sigma_L$ is a diagonal matrix that contains the singular values. 
Note that this SVD need only be computed once for each calibrated system. 
The pseudoinverse can then be efficiently pre-computed as $\Phi_L^{+} = V_L\Sigma_L^{-1}U_L^T$. 

When the matrices $\Phi_L$, $\Phi_R$ are not well-conditioned or are under-determined (e.g., when we have fewer measurements $M$ than the desired dimensionality of the scene $N$, as in compressive sensing \cite{candes2006stable,donoho2006compressed,baraniuk2007compressive}), some of the singular values are either very small or equal to zero. 
In these cases, the least-squares estimate $\widehat{X}_{\textrm{LS}}$ suffers from noise amplification. 
A simple approach to reduce noise amplification is to add an $\ell_2$ regularization term in the least-squares problem in \eqref{eq:LS}  
\begin{equation}\label{eq:tikLS}
\widehat{X}_{\textrm{Tik}} = \arg \min_X \; \|\Phi_L X \Phi_R^T-Y\|_2^2+\tau\|X\|_2^2,
\end{equation}
where $\tau>0$ is a regularization parameter. 
The solution of \eqref{eq:tikLS} can also be explicitly written using the SVD of $\Phi_L$ and $\Phi_R$ as we describe below. 

The solution of \eqref{eq:tikLS} can be computed by setting the gradient of the objective in \eqref{eq:tikLS} equal to zero and simplifying the resulting equation:
\begin{gather*}
\Phi_L^T(\Phi_LX\Phi_R^T-Y)\Phi_R+ \tau X  = 0 \\
\Phi_L^T\Phi_LX\Phi_R^T\Phi_R+ \tau X  = \Phi_L^TY\Phi_R.
\end{gather*}
Replacing $\Phi_L$ and $\Phi_R$ with their SVD decompositions yields
\[
V_L \Sigma_L^2 V_L^T X V_R \Sigma_R^2 V_R^T + \tau X = V_L\Sigma_LU_L^TYU_R\Sigma_R V_R^T.
\]
Multiplying both sides of the equation with $V_L^T$ from the left and $V_R$ from the right yields  
\begin{gather*} 
\Sigma_L^2 V_L^T X V_R \Sigma_R^2 + \tau  V_L^T X V_R = \Sigma_LU_L^TYU_R\Sigma_R.
\end{gather*}
Denote the diagonal entries of $\Sigma_L^2$ and $\Sigma_R^2$ using the vectors $\sigma_L$ and $\sigma_R$, respectively, to simplify the equations to
\begin{gather*}
V_L^T X V_R \odot (\sigma_L\sigma_R^T) + \tau V_L^T X V_R = \Sigma_LU_L^T Y U_R \Sigma_R \\
V_L^T X V_R \odot (\sigma_L\sigma_R^T+ \tau {\bf 11}^T)= \Sigma_LU_L^T Y U_R \Sigma_R \\
V_L^T X V_R = (\Sigma_LU_L^T Y U_R\Sigma_R)./({\sigma_L\sigma_R^T} + \tau  {\bf 1}{\bf 1}^T),
\end{gather*} 
where $A \odot B$ and $A./B$ denote element-wise multiplication and division of matrices $A$ and $B$, respectively. 
The solution of \eqref{eq:tikLS} can finally be written as 
\begin{equation}\label{eq:tikLS_solution}
\widehat{X}_\text{Tik} = V_L[(\Sigma_LU_L^T Y U_R\Sigma_R)./({\sigma_L\sigma_R^T} + \tau  {\bf 1}{\bf 1}^T)]V_R^T.
\end{equation}  
Thus, once the SVDs of $\Phi_L$ and $\Phi_R$ are computed and stored, reconstruction of an $N\times N$ image from $M\times M$ sensor measurements involves a fixed cost of two $M\times N$ matrix multiplications, two $N\times N$ matrix multiplications, and three $N\times N$ diagonal matrix multiplications.  

In many cases, exploiting the sparse or low-dimensional structure of the unknown image significantly enhances reconstruction performance.  
Natural images and videos exhibit a host of geometric properties, including sparse gradients and sparse coefficients in certain transform domains. Wavelet sparse models and total variation (TV) are widely used regularization methods for natural images \cite{mallat2008wavelet,rudin1992nonlinear}. 
By enforcing these geometric properties, we can suppress noise amplification as well as obtain unique solutions.
A pertinent example for image reconstruction is the sparse gradient model, which can be represented in the form of the following total-variation (TV) minimization problem: 
\begin{equation}
\widehat{X}_{\textrm{TV}} = \arg\min_X \|  \Phi_L X \Phi_R^T - Y\|^2 +  \lambda \| X \|_{\textrm{TV}}.
\label{eq:tv}
\end{equation}
The term $\| X \|_{\textrm{TV}}$ denotes the TV of the image $X$ given by the sum of magnitudes of the image gradients.
Given the scene $X$ as a 2D image, i.e., $X(u, v)$, we can define $G_u = D_u X$ and $G_v = D_v X$ as the spatial gradients of the image along the horizontal and vertical directions, respectively.
The total variation of the image is then defined as 
\[ 
\| X \|_{\textrm{TV}} = \sum_{u,v} \sqrt{ G_u(u, v)^2 +  G_v(u, v)^2 }. 
\]
Minimizing the TV as in (\ref{eq:tv}) produces images with sparse gradients.
The optimization problem (\ref{eq:tv}) is convex and can be efficiently solved using a variety of methods.
Many extensions and performance analyses are possible following the recently developed theory of compressive sensing.

In addition to simplifying the calibration task, separability of the coded mask also significantly reduces the computational burden of image reconstruction. Iterative methods for solving the optimization problems described above require the repeated application of the multiplexing matrix and its transpose. Continuing our numerical example from above, for a non-separable, dense mask, both of these operations would require on the order of $10^{12}$ multiplications and additions for mega-pixel images. With a separable mask, however, the application of the forward and transpose operators requires only on the order of $2\times10^9$ scalar multiplications and additions---a tremendous reduction in computational complexity.

\section{Experimental results}
We present results on two prototypes. The first uses a Silcon-based sensor to sense in visible wavelengths and the second uses an InGaAs sensor for sensing in short-wave infrared.

\subsection{Visible wavelength FlatCam prototype}

\begin{figure*}[t]
	\centering
	\includegraphics[width=0.75\textwidth, page=3]{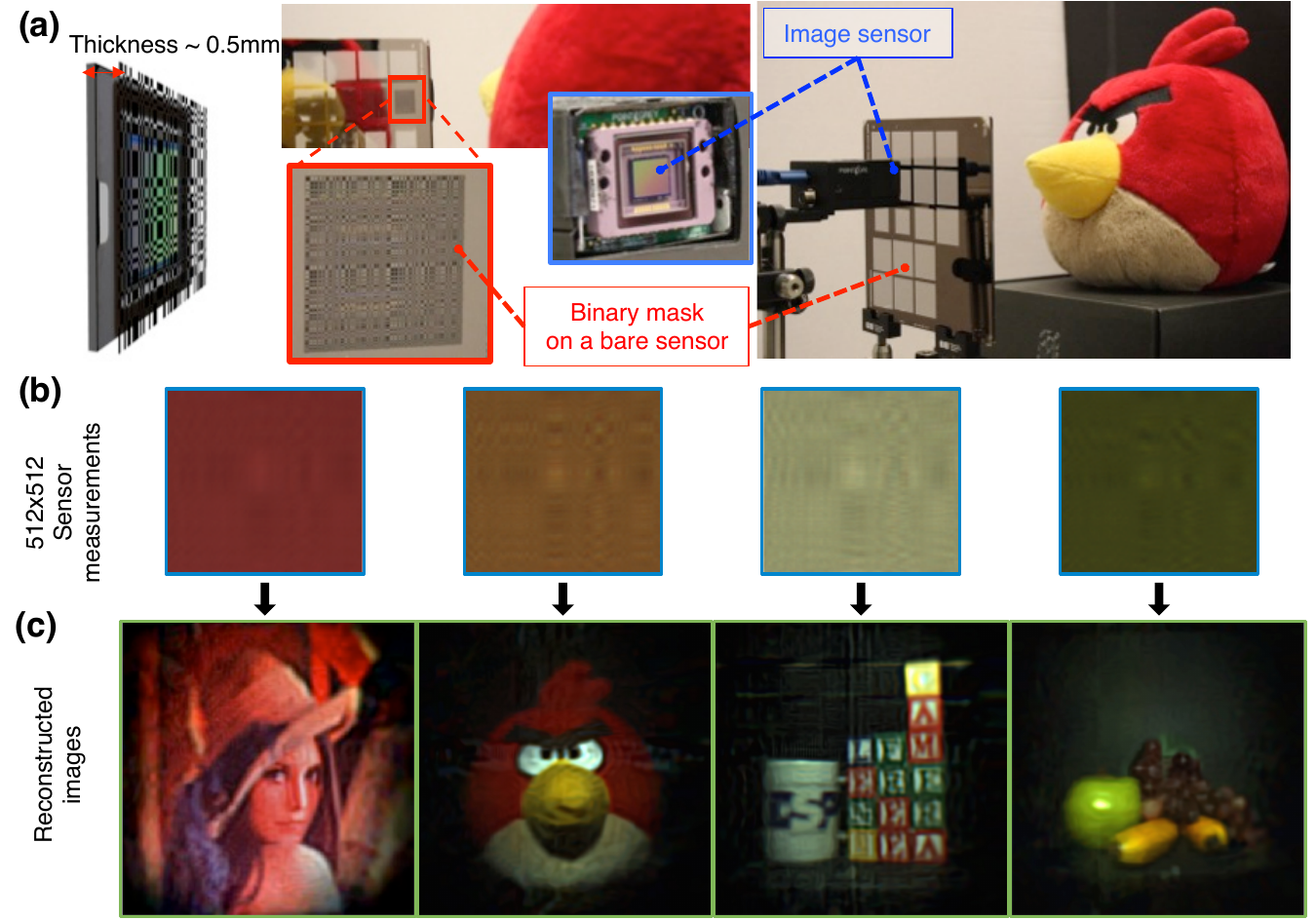}
	\caption{{Visible FlatCam prototype and results. \textbf{(a)}} Prototype consists of a Sony ICX285 sensor with a separable M-sequence mask placed approximately 0.5mm from the sensor surface. \textbf{(b)} The sensor measurements are different linear combinations of the light from different points in the scene. \textbf{(c)} Reconstructed $512\times 512$ color images by processing each color channel independently.}
	\label{fig:visible}
\end{figure*}

We built this FlatCam prototype as follows.

\noindent
\textbf{Image sensor:} We used a Sony ICX285 CCD color sensor that came inside a Point Grey Grasshopper 3 camera (model GS3-U3-14S5C-C). The sensor has $1036\times 1384$ pixels, each $6.45\mu$m wide, arranged in an RGB Bayer pattern. The physical size of the sensor array is approximately 6.7mm $\times$ 8.9mm.
\noindent
\textbf{Mask material:}  We used a custom-made chrome-on-quartz photomask that consists of a fused quartz plate, one side of which is covered with a pattern defined using a thin chrome film. The transparent regions of the mask transmit light, while the chrome film regions of the mask block light. 

\begin{figure}[!t]
	\centering
	\includegraphics[width=0.9\linewidth]{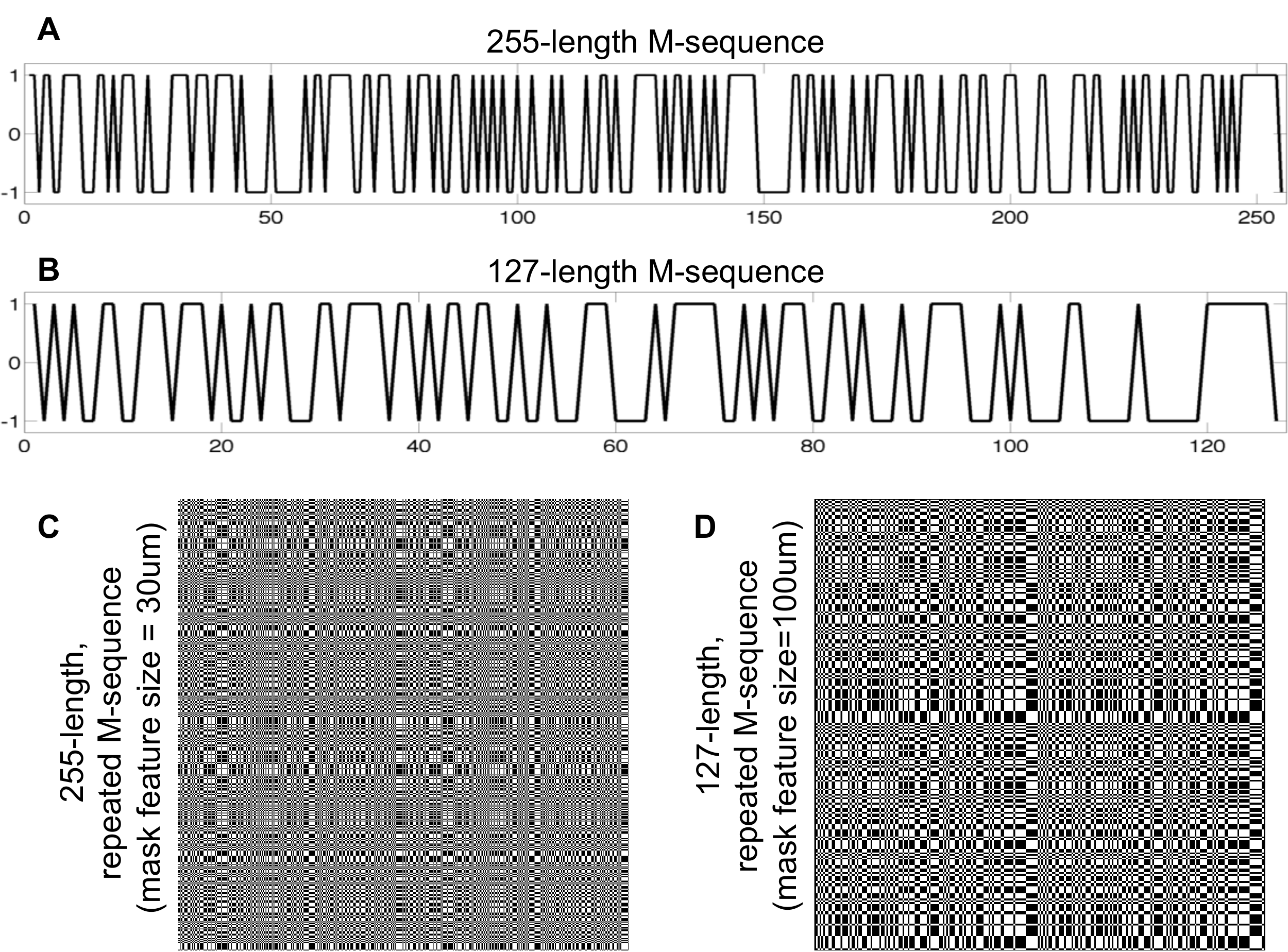}
	\caption{{ Masks used in both our visible and SWIR FlatCam prototypes.}  M-sequences with $\pm1$ entries that we used to create the binary masks for {\bf (a)} the visible camera and {\bf(b)} the SWIR camera.  Binary masks created from the M-sequences for {\bf (c)}~the visible camera and {\bf(d)} the SWIR camera.}
	\label{fig:masks}
\end{figure}

\noindent
\textbf{Mask pattern and resolution:} We created the binary mask pattern as follows. 
We first generated a length-255 M-sequence consisting of $\pm1$ entries. The actual 255-length M-sequence is shown in Fig.~\ref{fig:masks}. 
We repeated the M-sequence twice to create a 510-length sequence and computed the outer product with itself to create a $510 \times 510$ matrix. 
Since the resulting outer product consist of $\pm1$ entries, we replaced every $-1$ with a $0$ to create a binary matrix that is optically feasible. 
An image showing the final $510 \times 510$ mask pattern is shown in Fig.~\ref{fig:masks}.  
We printed a mask from the $510\times 510$ binary matrix such that each element corresponds to a $\Delta=30 \mu$m square box (transparent, if 1; opaque, if 0) on the printed mask.  
Images of the pattern that we used for the mask and the printed mask are presented in Fig.~\ref{fig:masks}.
The final printed mask is a square approximately 15.3mm on a side and covers the entire sensor area. 
Even though the binary mask is not separable as is, we can represent the sensor image using the separable system described in \eqref{eq:Y=PL X PR} by subtracting the row and column mean from the sensor images (see Sec.~\ref{sec:calibration} for details on calibration).

\noindent
\textbf{Mask placement:}  We opened the camera body to expose the sensor surface and placed the quartz mask on top of it using mechanical posts such that the mask touches the protective glass (hot mirror) on top of the sensor. Thus the distance between the mask and the sensor $d$ is determined by thickness of the glass, which for this sensor is $0.5$mm. 

\noindent
\textbf{Data readout and processing:} We adjusted the white balance of the sensor using Point Grey FlyCapture software and recorded images in 8-bit RGB format using suitable exposure and frame rate settings. In most of our experiments, the exposure time was fixed at 10ms, but we adjusted it according to the scene intensity to avoid excessively bright or dark sensor images. For the static scenes we averaged 20 sensor images to create a single set of measurements to be used for reconstruction. 

We reconstructed $512\times 512$ RGB images from our prototype using $512\times 512$ RGB sensor measurements. 
Since the sensor has $1086\times 1384$ pixels, we first cropped and uniformly subsampled the sensor image to create an effective $512\times 512$ color sensor image; then we subtracted the row and column means from that image. The resulting image corresponds to the measurements described by \eqref{eq:Y=PL X PR}, which we used to reconstruct the desired image $X$.
Some example sensor images and corresponding reconstruction results are shown in Fig.~\ref{fig:visible}. 
In these experiments, we solved an $\ell_2$-regularized least-squares problem in \eqref{eq:tikLS}, followed by BM3D denoising \cite{dabov2007image}. Solving the least-squares recovery problem for a single $512\times 512$ image using pre-computed SVD requires a fraction of a second on a standard laptop computer. 

\begin{figure*}[!t]
	\centering
	\includegraphics[width=0.95\textwidth,trim = 0mm 20mm 20mm 0mm, clip, page=7]{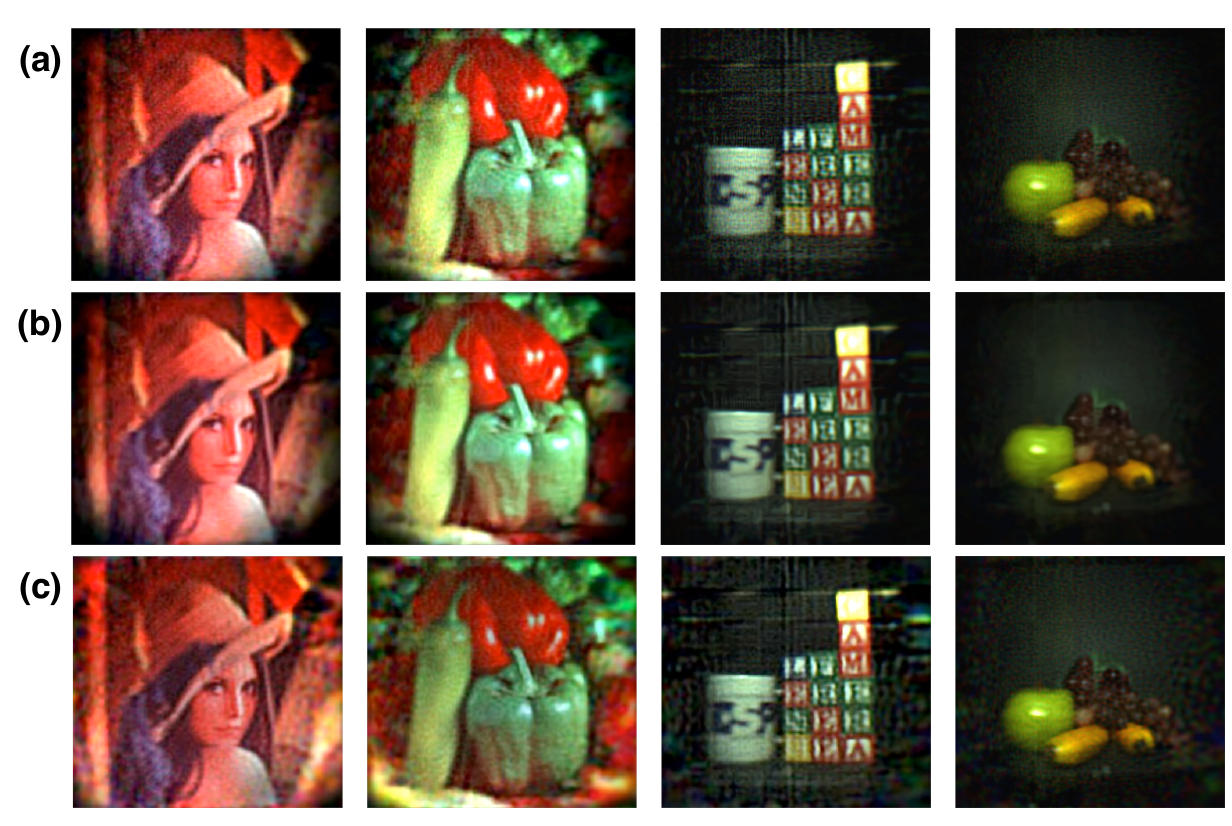}
	\caption{Images reconstructed at $512\times 512$ resolution using the visible FlatCam prototype and three different reconstruction methods. {\bf (a)} SVD-based solution of \eqref{eq:tikLS}; average computation time per image = 75ms. {\bf(b)} SVD/BM3D reconstruction; average computation time per image = 10s. {\bf(c)} Total variation (TV) based reconstruction; average computation time per image = 75s.}
	\label{fig:visible_S}
\end{figure*}

We present a comparison of three different methods for reconstructing static scenes in Fig.~\ref{fig:visible_S}. We used MATLAB for solving all the computational problems. For the results presented in Fig.~\ref{fig:visible_S}, we recorded sensor measurements while displaying  test images on an LCD monitor placed 28cm away from the camera and by placing various objects in front of the camera in ambient lighting.

We used three methods for reconstructing the scenes from the sensor measurements: 
\begin{enumerate}
	\item We computed and stored the SVD of $\Phi_L,\Phi_R$ and solved the $\ell_2$-regularized problem in \eqref{eq:tikLS} as described in \eqref{eq:tikLS_solution}. The average computation time for the reconstruction of a single $512\times 512$ image on a standard laptop was 75ms. The results of SVD-based reconstruction are presented in Fig.~\ref{fig:visible_S}\replace{B}{(a)}. The reconstructed images are slightly noisy, with details missing around the edges.
	\item To reduce the noise in our SVD-estimated images, we applied BM3D denoising to each reconstructed image. The results of SVD/BM3D reconstruction are presented in Fig.~\ref{fig:visible_S}\replace{C}{(b)}. The average computation time for BM3D denoising of a single image was 10s.
	\item To improve our results further, we reconstructed by solving the TV minimization problem \eqref{eq:tv}. \add{ The results of TV reconstruction are presented in Fig.~\ref{fig:visible_S}(c).} While, as expected, the TV method recovers more details around the edges, the overall reconstruction quality is not appreciably very different from SVD-based reconstruction. The computation time of TV, however, increases to 75s per image.
\end{enumerate}

\begin{figure*}[t]
	\centering
	\includegraphics[width=0.95\textwidth,trim = 0mm 40mm 0mm 0mm, clip, page=6]{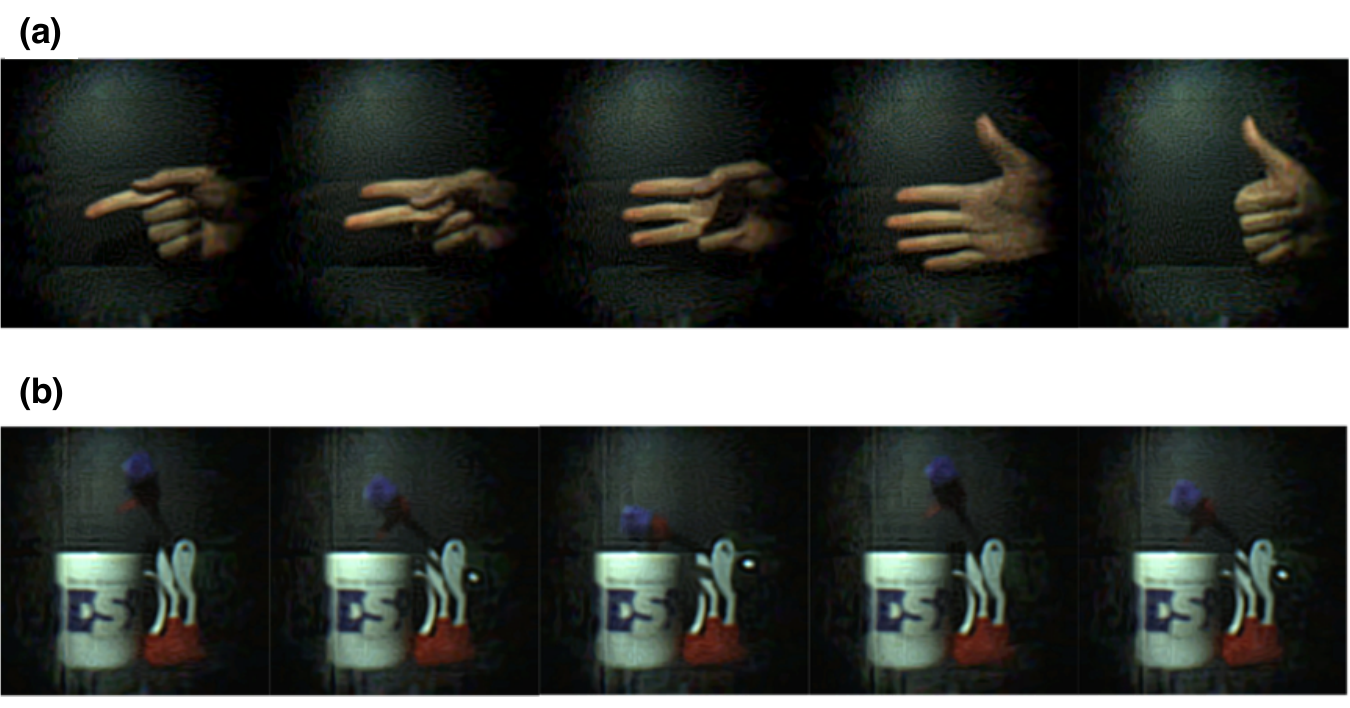}
	\caption{Dynamic scenes captured by a FlatCam at video rates and reconstructed at $512\times 512$ resolution. \textbf{(a)} Frames from the video of hand gestures captured at 30 frames per second. \textbf{(b)} Frames from the video of a toy bird captured at 10 frames per second.}
	\label{fig:videos}
\end{figure*}

To demonstrate the flexibility of our FlatCam design, we also captured and reconstructed dynamic scenes at typical video rates. We present selected frames\footnote{Complete videos are available at {\tt http://bit.ly/FlatCam}.} from two videos in Fig.~\ref{fig:videos}.
The images presented in Fig.~\ref{fig:videos}A are reconstructed frames from a video of a hand making counting gestures, recorded at 30 frames per second. 
The images presented in Fig.~\ref{fig:videos}B are reconstructed frames from a video of a toy bird dipping its head in water, recorded at 10 frames per second. In both cases, we reconstructed each video frame at $512\times 512$ pixel resolution by solving \eqref{eq:tikLS} using the SVD-based method described in \eqref{eq:tikLS_solution}, followed by BM3D denoising.

\subsection{SWIR FlatCam prototype} 
\begin{figure}[t]
	\centering
	\includegraphics[width=\linewidth ,trim = 0mm 35mm 35mm 0mm, clip, page=4]{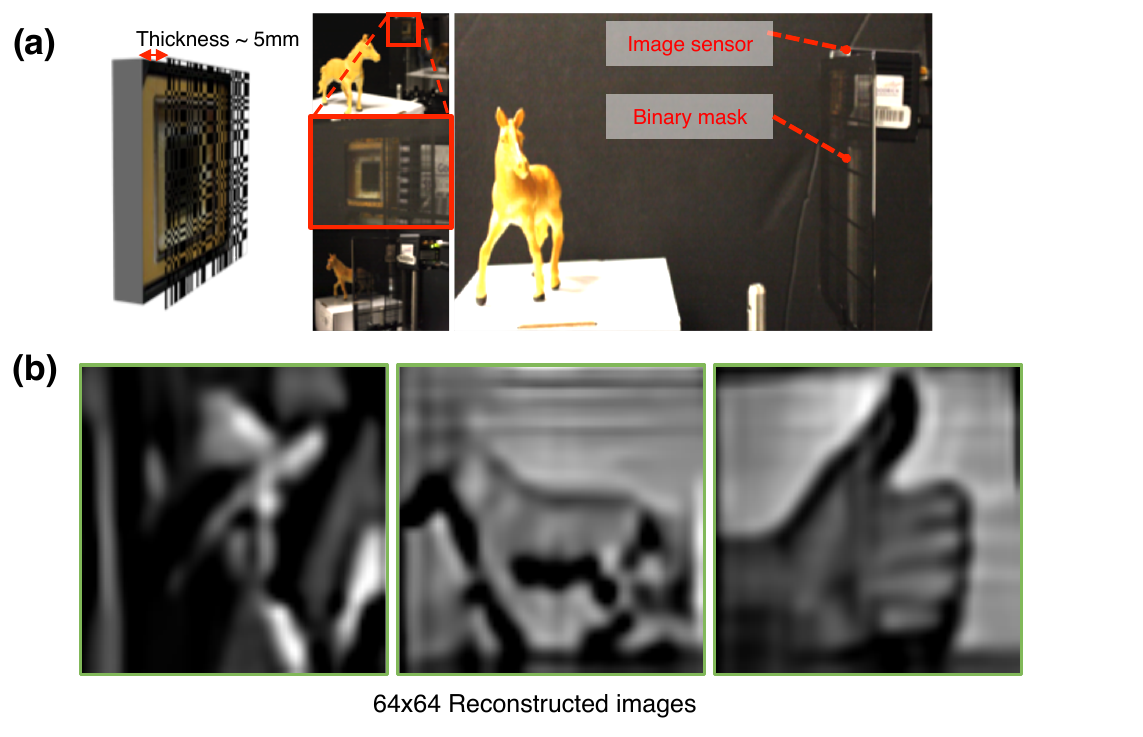}
	\caption{{Short wave infrared (SWIR) FlatCam prototype and results. \textbf{(a)}} Prototype consists of a Goodrich 320KTS-1.7RT sensor with a separable M-sequence mask placed approximately 5mm from the detector surface. \textbf{(b)} Reconstructed $64\times 64$ images. }
	\label{fig:swir}
\end{figure}

This FlatCam prototype consists of a Goodrich 320KTS-1.7RT InGaAs sensor with a binary separable M-sequence mask placed at distance $d=5$mm. The sensor-mask distance is large in this prototype because of the protective casing on top of the sensor. We used a feature size of $\Delta=100\mu$m for the mask, which was constructed using the same photomask process as for the visible camera. The sensor has $256\times 300$  pixels, each of size $w=25\mu$m, but because of the large sensor-to-mask distance and mask feature size, the effective system resolution is limited. Therefore, we binned $4\times 4$ pixels on the sensor (and cropped a square region of the sensor) to produce sensor measurements of effective size $64\times 64$. We reconstructed images with the same $64\times 64$  resolution; example results are shown in Fig.~\ref{fig:swir}. 

\section{Discussions and Conclusions}
The mask-based, lens-free FlatCam design proposed here can have a significant impact in an important emerging area of imaging, since high-performance, broad-spectrum cameras can be monolithically fabricated instead of requiring cumbersome post-fabrication assembly. The thin form factor and low cost of lens-free cameras makes them ideally suited for many applications in surveillance, large surface cameras, flexible or foldable cameras, disaster recovery, and beyond, where cameras are either disposable resources or integrated in flat or flexible surfaces and therefore have to satisfy strict thickness constraints. Emerging applications like wearable devices, internet-of-things, and in-vivo imaging could also benefit from the FlatCam approach.

\subsection{Advantages of FlatCam}
We make key changes in our FlatCam design to move away from the cube-like form-factor of traditional lens-based and coded aperture cameras while retaining their high light collection abilities. We move the coded mask extremely close to the image sensor, which results in a thin, flat camera. We use a binary mask pattern with 50\% transparent features, which, when combined with the large surface area sensor, enables large light collection capabilities. We use a separable mask pattern, similar to the prior work in coded aperture imaging \cite{deweert2015lensless}, which enables simpler calibration and reconstruction. The result is a radically different form factor from previous camera designs that can enable integration of FlatCams into large surfaces and flexible materials such as wallpaper and clothes that require thin, flat, and lightweight materials \cite{koppens2014photodetectors}.

\textit{Flat form factor.}
The flatness of a camera system can be measured by its thickness-to-width ratio (TWR). The form factor of most cameras, including  pinhole and lens-based cameras, conventional coded-aperture systems \cite{fenimore1978coded}, and miniature diffraction grating-based systems \cite{gill2013lensless}, is cube-like; that is, the thickness of the device is of the same order of magnitude as the sensor width, resulting in $\text{TWR}\approx 1$. Cube-like camera systems suffer from a significant limitation: if we reduce the thickness of the camera by an order of magnitude while preserving its TWR, then the area of the sensor drops by two order of magnitude. This results in a two orders of magnitude reduction in light collection ability. In contrast, FlatCams are endowed with flat form factors; by design, the thickness of the device is an order of magnitude smaller than the sensor width. Thus, for a given thickness constraint, a FlatCam can utilize a large sensing surface for light collection. In our visible FlatCam prototype, for example, the sensor-to-mask distance is 0.5mm, while the sensor width is about 6.7mm, resulting in $\text{TWR}\approx 0.075$. While on-chip lensless microscopes can also achieve such low TWRs, such systems require complete control of the illumination and the subject to be less than 1 mm from the camera \cite{greenbaum2012imaging}. We are unaware of any other far-field imaging system that has a comparable TWR of the FlatCam while providing reasonable light capture and imaging resolution.

\textit{High light collection.}
The light collection ability of an imaging system depends on two factors: its sensor area and the square of its numerical aperture. Conventional sensor pixels typically have an angular response of 40--60 degrees, which is referred to as the sensor’s chief ray angle (CRA). The total amount of light that can be sensed by a sensor is often limited by the CRA, which in turn determines the maximum allowable numerical aperture of the system. Specifically, whether we consider the best lens-based camera, or even a fully exposed sensor, the cone of light that can enter a pixel is determined by the CRA. 

Consider an imaging system with a strict constraint on the device thickness $T_\text{max}$. The light collection $L$ of such an imaging device can be described as $L\propto W^2 N_A^2$, where $W$ denotes the width of the (square) sensor and $N_A$ denotes the numerical aperture. Since $W_\text{max}=T_\text{max}/\text{TWR}$, we have $L\propto  W^2 N_A^2  \le (N_A  T_\text{max}/\text{TWR})^2$. Thus, given a thickness constraint $T_\text{max}$, the light collection of an imaging system is directly proportional to the square of the numerical aperture and inversely proportional to the square of its TWR. Thus, smaller TWR leads to better light collection.

The numerical aperture of our prototype FlatCams is limited by the CRA of the sensors. Moreover, half of the features in our mask are opaque and block one half of the light that would have otherwise entered the sensor. Realizing that the numerical aperture of such a FlatCam is reduced only by a factor of $\sqrt{2}$ compared to an open aperture, yet its TWR is reduced by an order of magnitude leads to the conclusion that a FlatCam collects approximately two orders of magnitude more light than a cube-like miniature camera of the same thickness.

\subsection{Limitations of FlatCam}
FlatCam is a radical departure from centuries of research and development in lens-based cameras, and as such this radical departure has its own limitations.

\textit{Achievable image/angular resolution.}
Our current prototypes have low spatial resolution which is attributed to two factors.
First, it is well known that angular resolution of pinhole cameras and coded aperture cameras decreases when the mask is moved closer to the sensor \cite{brady2009optical}.
This results in an implicit tradeoff between the achievable thickness and the achievable resolution.
Second, the image recorded on the image sensor in a FlatCam is a linear combination of the scene radiance, where the multiplexing matrix is controlled by the mask pattern and distance between mask and sensor.
This means that recovering the scene from sensor measurements requires demultiplexing. Noise amplification is an unfortunate outcome of any linear demultiplexing based system. While the magnitude of this noise amplification can be controlled by careful design of the mask patterns, they cannot be completely eliminated in FlatCam. In addition, the singular values of the linear system are such that the noise amplification for higher spatial frequencies is larger, which consequently limits the spatial resolution of the recovered image. We are currently working on several techniques to improve the spatial resolution of the recovered images.

\textit{Direct-view and real-time operation.}
In traditional lens-based cameras, the image sensed by the image sensor is the photograph of the scene. In FlatCam, a computational algorithm is required to convert the sensor measurements into a photograph of the scene. This results in a time-lag between the sensor acquisition and the image display, a time-lag that depends on processing time. Currently, our SVD-based reconstruction operates at near real-time (about 10 fps) resulting in about a 100 ms delay between capture and display. While this may be acceptable for certain applications, there are many other applications such as augmented reality and virtual reality, where such delays are unacceptable. Order of magnitude improvements in processing times are required before FlatCam becomes amenable to such applications.

\section*{Acknowledgments}

This work was supported by NSF grants IIS--1116718, CCF--1117939, and CCF--1527501 and ARO MURI W911NF0910383.

\ifCLASSOPTIONcaptionsoff
  \newpage
\fi



\bibliographystyle{IEEEtran}
\bibliography{flatcam_arxiv.bib}

\begin{thebibliography}{10}
\providecommand{\url}[1]{#1}
\csname url@samestyle\endcsname
\providecommand{\newblock}{\relax}
\providecommand{\bibinfo}[2]{#2}
\providecommand{\BIBentrySTDinterwordspacing}{\spaceskip=0pt\relax}
\providecommand{\BIBentryALTinterwordstretchfactor}{4}
\providecommand{\BIBentryALTinterwordspacing}{\spaceskip=\fontdimen2\font plus
\BIBentryALTinterwordstretchfactor\fontdimen3\font minus
  \fontdimen4\font\relax}
\providecommand{\BIBforeignlanguage}[2]{{%
\expandafter\ifx\csname l@#1\endcsname\relax
\typeout{** WARNING: IEEEtran.bst: No hyphenation pattern has been}%
\typeout{** loaded for the language `#1'. Using the pattern for}%
\typeout{** the default language instead.}%
\else
\language=\csname l@#1\endcsname
\fi
#2}}
\providecommand{\BIBdecl}{\relax}
\BIBdecl

\bibitem{dicke1968scatter}
R.~Dicke, ``Scatter-hole cameras for x-rays and gamma rays,'' \emph{The
  Astrophysical Journal}, vol. 153, p. L101, 1968.

\bibitem{fenimore1978coded}
E.~Fenimore and T.~Cannon, ``Coded aperture imaging with uniformly redundant
  arrays,'' \emph{Applied optics}, vol.~17, no.~3, pp. 337--347, 1978.

\bibitem{gottesman1989new}
S.~R. Gottesman and E.~Fenimore, ``New family of binary arrays for coded
  aperture imaging,'' \emph{Applied optics}, vol.~28, no.~20, pp. 4344--4352,
  1989.

\bibitem{cannon1980coded}
T.~Cannon and E.~Fenimore, ``Coded aperture imaging: Many holes make light
  work,'' \emph{Optical Engineering}, vol.~19, no.~3, pp. 193--283, 1980.

\bibitem{durrant1999application}
P.~Durrant, M.~Dallimore, I.~Jupp, and D.~Ramsden, ``The application of pinhole
  and coded aperture imaging in the nuclear environment,'' \emph{Nuclear
  Instruments and Methods in Physics Research Section A: Accelerators,
  Spectrometers, Detectors and Associated Equipment}, vol. 422, no.~1, pp.
  667--671, 1999.

\bibitem{dragoi2010cmos}
V.~Dragoi, A.~Filbert, S.~Zhu, and G.~Mittendorfer, ``Cmos wafer bonding for
  back-side illuminated image sensors fabrication,'' in \emph{2010 11th
  International Conference on Electronic Packaging Technology \& High Density
  Packaging}, 2010, pp. 27--30.

\bibitem{brady2009optical}
D.~J. Brady, \emph{Optical imaging and spectroscopy}.\hskip 1em plus 0.5em
  minus 0.4em\relax John Wiley \& Sons, 2009.

\bibitem{zomet2006lensless}
A.~Zomet and S.~K. Nayar, ``Lensless imaging with a controllable aperture,'' in
  \emph{IEEE Computer Society Conference on Computer Vision and Pattern
  Recognition}, vol.~1, 2006, pp. 339--346.

\bibitem{huang2013lensless}
G.~Huang, H.~Jiang, K.~Matthews, and P.~Wilford, ``Lensless imaging by
  compressive sensing,'' in \emph{20th IEEE International Conference on Image
  Processing}, 2013, pp. 2101--2105.

\bibitem{deweert2015lensless}
M.~J. DeWeert and B.~P. Farm, ``Lensless coded-aperture imaging with separable
  doubly-toeplitz masks,'' \emph{Optical Engineering}, vol.~54, no.~2, pp.
  023\,102--023\,102, 2015.

\bibitem{levin2007image}
A.~Levin, R.~Fergus, F.~Durand, and W.~T. Freeman, ``Image and depth from a
  conventional camera with a coded aperture,'' in \emph{ACM Transactions on
  Graphics (TOG)}, vol.~26, no.~3.\hskip 1em plus 0.5em minus 0.4em\relax ACM,
  2007, p.~70.

\bibitem{liu_PCeV2013}
D.~Liu, J.~Gu, Y.~Hitomi, M.~Gupta, T.~Mitsunaga, and S.~Nayar, ``{E}fficient
  {S}pace-{T}ime {S}ampling with {P}ixel-wise {C}oded {E}xposure for {H}igh
  {S}peed {I}maging,'' \emph{IEEE Transactions on Pattern Analysis and Machine
  Intelligence}, vol.~99, p.~1, 2013.

\bibitem{veeraraghavan2007dappled}
A.~Veeraraghavan, R.~Raskar, A.~Agrawal, A.~Mohan, and J.~Tumblin, ``Dappled
  photography: Mask enhanced cameras for heterodyned light fields and coded
  aperture refocusing,'' \emph{ACM Trans. Graph.}, vol.~26, no.~3, p.~69, 2007.

\bibitem{marwah2013compressive}
K.~Marwah, G.~Wetzstein, Y.~Bando, and R.~Raskar, ``Compressive light field
  photography using overcomplete dictionaries and optimized projections,''
  \emph{ACM Transactions on Graphics (TOG)}, vol.~32, no.~4, p.~46, 2013.

\bibitem{candes2006stable}
E.~J. Candes, J.~K. Romberg, and T.~Tao, ``Stable signal recovery from
  incomplete and inaccurate measurements,'' \emph{Communications on pure and
  applied mathematics}, vol.~59, no.~8, pp. 1207--1223, 2006.

\bibitem{donoho2006compressed}
D.~L. Donoho, ``Compressed sensing,'' \emph{IEEE Transactions on Information
  Theory}, vol.~52, no.~4, pp. 1289--1306, 2006.

\bibitem{baraniuk2007compressive}
R.~G. Baraniuk, ``Compressive sensing,'' \emph{IEEE signal processing
  magazine}, vol.~24, no.~4, 2007.

\bibitem{marcia2008compressive}
R.~F. Marcia and R.~M. Willett, ``Compressive coded aperture superresolution
  image reconstruction,'' in \emph{IEEE International Conference on Acoustics,
  Speech and Signal Processing (ICASSP)}, 2008, pp. 833--836.

\bibitem{wagadarikar2008single}
A.~Wagadarikar, R.~John, R.~Willett, and D.~Brady, ``Single disperser design
  for coded aperture snapshot spectral imaging,'' \emph{Applied optics},
  vol.~47, no.~10, pp. B44--B51, 2008.

\bibitem{llull2013coded}
P.~Llull, X.~Liao, X.~Yuan, J.~Yang, D.~Kittle, L.~Carin, G.~Sapiro, and D.~J.
  Brady, ``Coded aperture compressive temporal imaging,'' \emph{Optics
  express}, vol.~21, no.~9, pp. 10\,526--10\,545, 2013.

\bibitem{tanida2001thin}
J.~Tanida, T.~Kumagai, K.~Yamada, S.~Miyatake, K.~Ishida, T.~Morimoto,
  N.~Kondou, D.~Miyazaki, and Y.~Ichioka, ``{Thin observation module by bound
  optics (TOMBO): concept and experimental verification},'' \emph{Applied
  optics}, vol.~40, no.~11, pp. 1806--1813, 2001.

\bibitem{shankar2008thin}
M.~Shankar, R.~Willett, N.~Pitsianis, T.~Schulz, R.~Gibbons, R.~Te~Kolste,
  J.~Carriere, C.~Chen, D.~Prather, and D.~Brady, ``Thin infrared imaging
  systems through multichannel sampling,'' \emph{Applied optics}, vol.~47,
  no.~10, pp. B1--B10, 2008.

\bibitem{bruckner2010thin}
A.~Br{\"u}ckner, J.~Duparr{\'e}, R.~Leitel, P.~Dannberg, A.~Br{\"a}uer, and
  A.~T{\"u}nnermann, ``Thin wafer-level camera lenses inspired by insect
  compound eyes,'' \emph{Optics Express}, vol.~18, no.~24, pp.
  24\,379--24\,394, 2010.

\bibitem{venkataraman2013picam}
K.~Venkataraman, D.~Lelescu, J.~Duparr{\'e}, A.~McMahon, G.~Molina,
  P.~Chatterjee, R.~Mullis, and S.~Nayar, ``Picam: An ultra-thin high
  performance monolithic camera array,'' \emph{ACM Transactions on Graphics
  (TOG)}, vol.~32, no.~6, p. 166, 2013.

\bibitem{tremblay2007ultrathin}
E.~J. Tremblay, R.~A. Stack, R.~L. Morrison, and J.~E. Ford, ``Ultrathin
  cameras using annular folded optics,'' \emph{Applied optics}, vol.~46, no.~4,
  pp. 463--471, 2007.

\bibitem{wang2009angle}
A.~Wang, P.~Gill, and A.~Molnar, ``Angle sensitive pixels in cmos for lensless
  3d imaging,'' in \emph{IEEE Custom Integrated Circuits Conference}, 2009, pp.
  371--374.

\bibitem{gill2011microscale}
P.~R. Gill, C.~Lee, D.-G. Lee, A.~Wang, and A.~Molnar, ``A microscale camera
  using direct fourier-domain scene capture,'' \emph{Optics letters}, vol.~36,
  no.~15, pp. 2949--2951, 2011.

\bibitem{gill2013lensless}
P.~R. Gill and D.~G. Stork, ``Lensless ultra-miniature imagers using
  odd-symmetry spiral phase gratings,'' in \emph{Computational Optical Sensing
  and Imaging}.\hskip 1em plus 0.5em minus 0.4em\relax Optical Society of
  America, 2013, pp. CW4C--3.

\bibitem{stork2013sensorcomm}
D.~Stork and P.~Gill, ``Lensless ultra-miniature cmos computational imagers and
  sensors,'' in \emph{International Conference on Sensor Technologies and
  Applications}, 2013, pp. 186--190.

\bibitem{greenbaum2012imaging}
A.~Greenbaum, W.~Luo, T.-W. Su, Z.~G{\"o}r{\"o}cs, L.~Xue, S.~O. Isikman, A.~F.
  Coskun, O.~Mudanyali, and A.~Ozcan, ``Imaging without lenses: Achievements
  and remaining challenges of wide-field on-chip microscopy,'' \emph{Nature
  methods}, vol.~9, no.~9, pp. 889--895, 2012.

\bibitem{greenbaum2014wide}
A.~Greenbaum, Y.~Zhang, A.~Feizi, P.-L. Chung, W.~Luo, S.~R. Kandukuri, and
  A.~Ozcan, ``Wide-field computational imaging of pathology slides using
  lens-free on-chip microscopy,'' \emph{Science translational medicine},
  vol.~6, no. 267, pp. 267ra175--267ra175, 2014.

\bibitem{macwilliams1976pseudo}
F.~J. MacWilliams and N.~J. Sloane, ``Pseudo-random sequences and arrays,''
  \emph{Proceedings of the IEEE}, vol.~64, no.~12, pp. 1715--1729, 1976.

\bibitem{busboom1998uniformly}
A.~Busboom, H.~Elders-Boll, and H.~Schotten, ``Uniformly redundant arrays,''
  \emph{Experimental Astronomy}, vol.~8, no.~2, pp. 97--123, 1998.

\bibitem{golomb1982shift}
S.~W. Golomb, \emph{Shift register sequences}.\hskip 1em plus 0.5em minus
  0.4em\relax Aegean Park Press, 1982.

\bibitem{fenimore1981fast}
E.~Fenimore and G.~Weston, ``Fast delta hadamard transform,'' \emph{Applied
  optics}, vol.~20, no.~17, pp. 3058--3067, 1981.

\bibitem{ding2015complementary}
J.~Ding, M.~Noshad, and V.~Tarokh, ``Complementary lattice arrays for coded
  aperture imaging,'' \emph{arXiv preprint arXiv:1506.02160}, 2015.

\bibitem{mallat2008wavelet}
S.~Mallat, \emph{A wavelet tour of signal processing: the sparse way}.\hskip
  1em plus 0.5em minus 0.4em\relax Academic press, 2008.

\bibitem{rudin1992nonlinear}
L.~I. Rudin, S.~Osher, and E.~Fatemi, ``Nonlinear total variation based noise
  removal algorithms,'' \emph{Physica D: Nonlinear Phenomena}, vol.~60, no.~1,
  pp. 259--268, 1992.

\bibitem{dabov2007image}
K.~Dabov, A.~Foi, V.~Katkovnik, and K.~Egiazarian, ``Image denoising by sparse
  3-d transform-domain collaborative filtering,'' \emph{IEEE Transactions on
  Image Processing}, vol.~16, no.~8, pp. 2080--2095, 2007.

\bibitem{koppens2014photodetectors}
F.~Koppens, T.~Mueller, P.~Avouris, A.~Ferrari, M.~Vitiello, and M.~Polini,
  ``Photodetectors based on graphene, other two-dimensional materials and
  hybrid systems,'' \emph{Nature nanotechnology}, vol.~9, no.~10, pp. 780--793,
  2014.

\end{thebibliography}

\end{document}